\pdfoutput=1
\documentclass[10pt,twocolumn,letterpaper]{article}

\usepackage{cvpr}
\usepackage{times}
\usepackage{epsfig}
\usepackage{graphicx}
\usepackage{amsmath}
\usepackage{amssymb}
\usepackage{array}
\usepackage{subcaption}
\usepackage{multirow}

\usepackage[pagebackref=true,breaklinks=true,letterpaper=true,colorlinks,bookmarks=false]{hyperref}

\cvprfinalcopy 

\def\cvprPaperID{6426} 

\ifcvprfinal\fi
\begin{document}

\title{\LARGE \bf
Predictive and Semantic Layout Estimation for Robotic Applications in Manhattan Worlds
}

\author{Armon Shariati, Bernd Pfrommer, and Camillo J. Taylor\\
University of Pennsylvania\\
Philadelphia PA, 19104\\
{\tt\small \{armon, pfrommer, cjtaylor\}@seas.upenn.edu}
}

\maketitle

\begin{abstract}
This paper describes an approach to automatically extracting floor plans from the kinds of incomplete measurements that could be acquired by an autonomous mobile robot. The approach proceeds by reasoning about extended structural layout surfaces which are automatically extracted from the available data. The scheme can be run in an online manner to build water tight representations of the environment. The system effectively speculates about room boundaries and free space regions which provides useful guidance to subsequent motion planning systems. Experimental results are presented on multiple data sets.
\end{abstract}

\section{Introduction}

\begin{figure*}[!ht]
    \centering
    \begin{subfigure}{0.3\textwidth}
        \centering
        \includegraphics[width=0.9\textwidth]{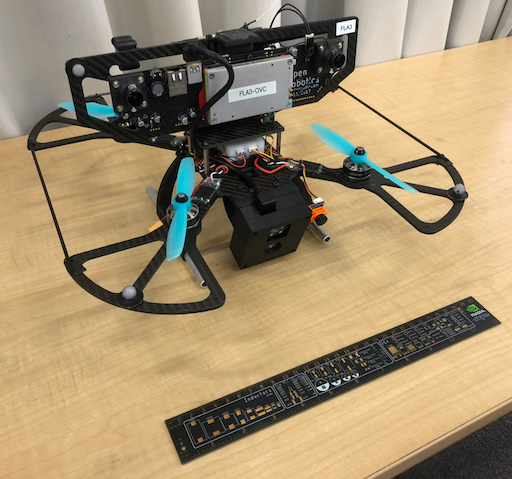}
        \caption{Quadrotor w/ Sensor Suite}
        \label{fig:falcon}
    \end{subfigure}
    \begin{subfigure}{0.39\textwidth}
        \centering
        \includegraphics[width=\textwidth,clip,trim={0.7in 0.5in 0.4in 1in},keepaspectratio]{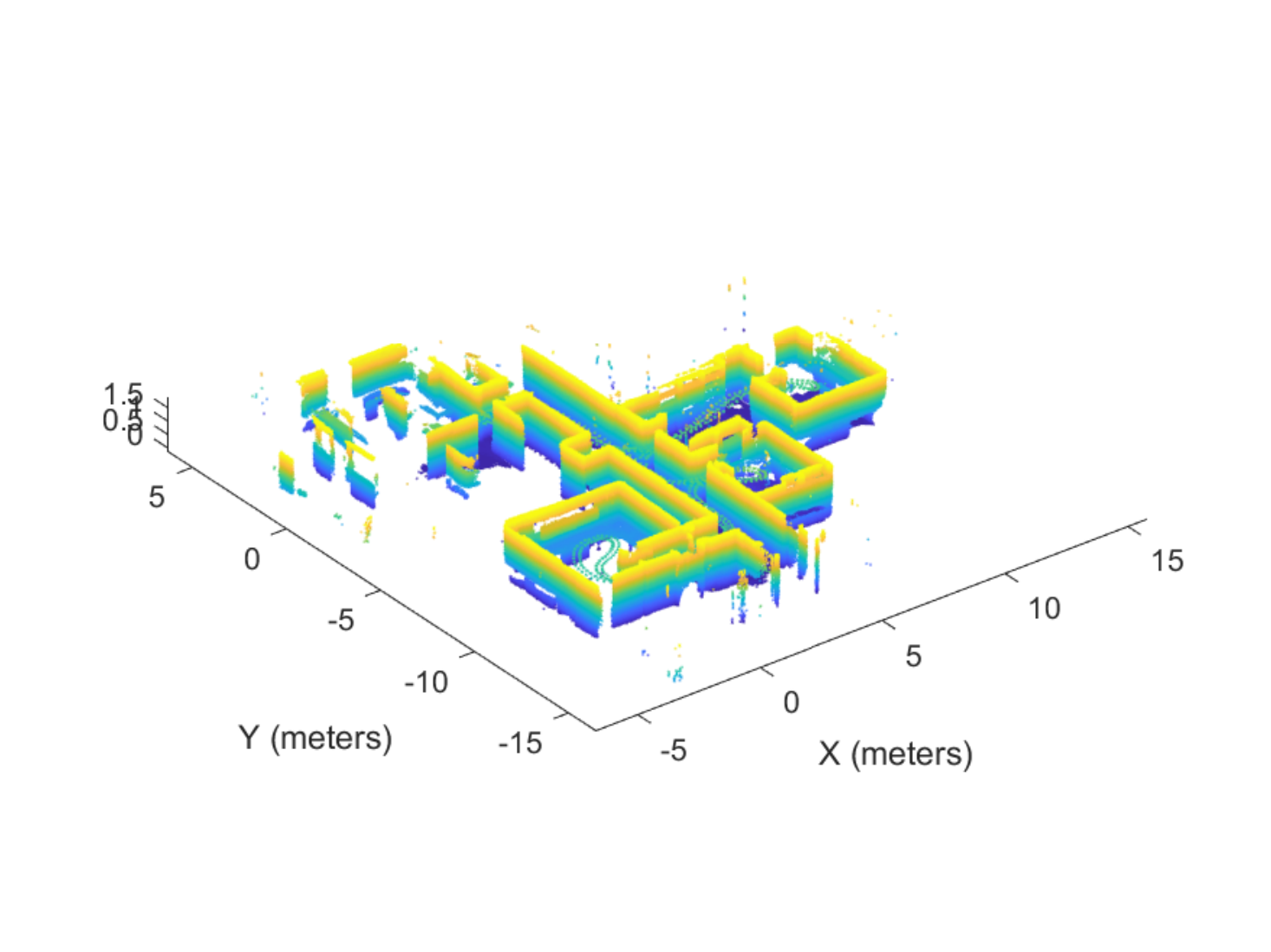}
        \caption{3D Point Cloud Reconstruction}
    \end{subfigure}
    \begin{subfigure}{0.3\textwidth}
        \centering
        \includegraphics[width=0.9\textwidth,clip,trim={0.8in 0in 1.2in 0.4in}]{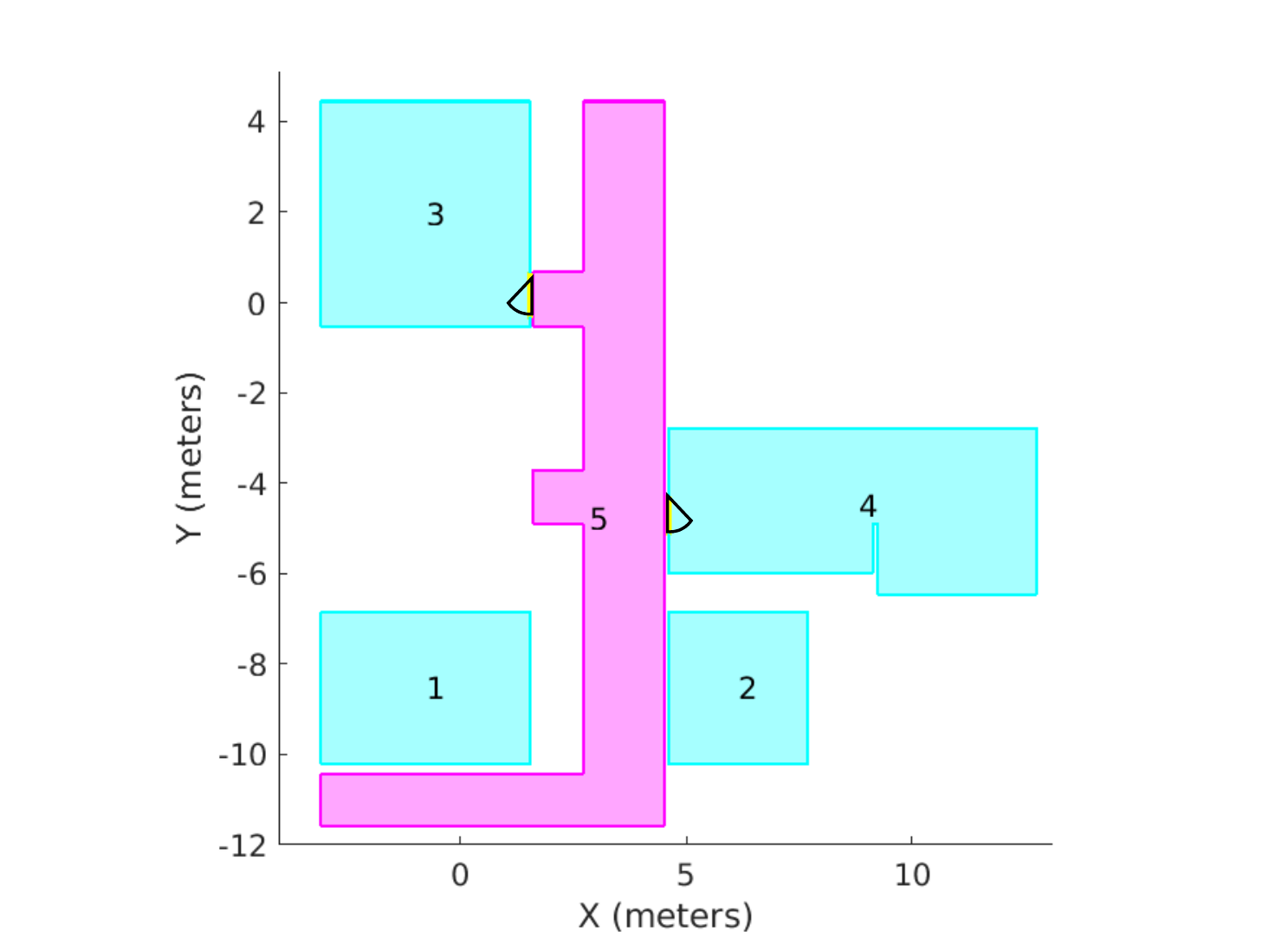}
        \caption{Semantic Floor Plan}
    \end{subfigure}
    \caption{The goal of this work is to be able to automatically construct semantic layouts of indoor spaces based on the kinds of data that could be acquired from an autonomous robot like the one shown in (a). This system is equipped with a pair of stereo cameras, an IMU and a PMD depth camera. (b) Shows a small portion of the 3D point cloud that we can acquire by integrating information from the robots sensors (c) Shows the abstracted floor plan distilled from the 3D measurements that are acquired as the sensor suite is moved through the scene.}
    \label{fig:intro}
\end{figure*}

In this paper we present an algorithm which is designed to generate semantic floor plans for an autonomous agent, such as the one shown in Figure \ref{fig:intro}, that must navigate through previously unknown indoor environments. In many such applications it is neither necessary nor desirable for the robot to view every corner or crevasse of the environment. Humans moving through such a scene are able to quickly apprehend the overall structure and make reasonable inferences about the structure of the scene even with incomplete information gleaned from a few scans and we would like to endow our robots with similar capabilities.

One of our goals is to produce floor plans that are useful in solving motion planning problems. As such our method is designed to be computationally efficient, to provide robust speculation as to the presence of free space and boundaries in regions of the environment which have not yet been observed, and to provide a semantic context for higher level planning tasks.

Many current layout estimation pipelines focus on scene understanding from a single image, or focus on generating models only after the entire space has been observed. Our approach fuses information from multiple vantage points and can be run in an online manner to provide predictions as the robot moves.

\section{Related Work}

\subsection{Single-View Layout Estimation}

A number of methods have been proposed that use geometric methods to infer scene layout from a single RGB image including
 \cite{Hedau2012,Schwing2013,Gupta2010a,Guo2013a}. These methods typically focus on detecting vanishing points within a Manhattan frame of reference in order to estimate layout through scene occlusion. Cowley \etal \cite{Taylor2012} present a related single view approach that uses depth images rather than RGB images.

More recently, several learning based approaches to the single view variant of the problem have been developed. Lee \etal \cite{Lee2017} use CNN features, instead of the more traditional handcrafted ones, in order to detect vanishing points and, in turn, scene layout. Methods such as \cite{Tulsiani2017a} and \cite{Song2017} can produce quite flexible factored representations of the completed scene, however, neither produce the water-tight floor plan models which are desirable in robotic applications. Finally, while \cite{Zou2018a} and \cite{Fernandez-Labrador2018a} are able to produce such floor plans, both rely on panoramic images, which may be difficult to obtain on a robotic platform.

Ultimately, despite the success and apparent efficacy of these systems, most fail to present a way in which additional information from multiple views can be incorporated into the information processing pipeline to model extended environments.

\subsection{Multi-View Layout Estimation}

The class of multi-view layout estimation techniques is comprised of solutions arising predominantly from three different problem domains. 

The first domain is robotics, where the goal is often to distill a semantically meaningful representation of an environment from sequential image data. Like this paper, these works often target real-time planning and navigation applications. Among the first of these approaches are \cite{Flint2011} and \cite{Tsai2011} which use a Bayesian filtering framework to update a model of the scene based on observations provided by some state-estimation subsystem. Although we may share many of the same goals and high-level approaches as some of these authors, our work expands upon their insights with a focus on prediction and model utility. A learning approach designed for a similar end can be found in \cite{Liu2018a}, where the authors use RGBD information to produce semantic floor plans, yet they do not emphasize the need for speculation in their system as we do.

The second subset of techniques is automated floor plan generation. Okorn \etal \cite{Okorn2010} focuses on generating accurate 2D floor plans by extracting and reasoning about complete wall segments from high-fidelity point clouds. In \cite{Ochmann2016} and \cite{Mura2016}, the authors utilize a 3D piecewise-planar representation and focus on volumetric reasoning of space. These techniques are appealing as they can produce more flexible models of indoor space than those permitted by making a Manhattan world assumption, however, they also require that all of the data be of a certain quality and available at once.

The third category is that of large-scale indoor modeling, where the work most closely related to ours may be found. 
The approach of Xiao \etal \cite{Xiao2014} is similar to ours in that the authors leverage aspects of the Manhattan world assumption and use rectangular primitives to model free space. Our approach differs in that we build our candidate regions using infinitely extended structural planes that are automatically detected as opposed to finte wall extents. Furthermore, our online approach is specifically designed to suggest unexplored regions that could be freespace which can be contrasted with the batch oriented approach in \cite{Xiao2014} which is discouraged from including unexplored regions in its volumetric model. In Armeni \etal \cite{Armeni2016}, the authors present their work focused on large-scale semantic parsing, which demonstrates how larger point clouds, such as those of a building, can be parsed into semantically meaningful components such as rooms, hallways, etc. However, despite providing a useful labeling of the points, their model of the environment is still, ultimately, a point cloud instead of a floor plan. 
\section{Technical Approach}

\begin{figure*}
    \centering
    \begin{subfigure}{0.3\textwidth}
        \includegraphics[width=0.9\textwidth,clip,trim={1.6in 0in 1.8in 0.3in}]{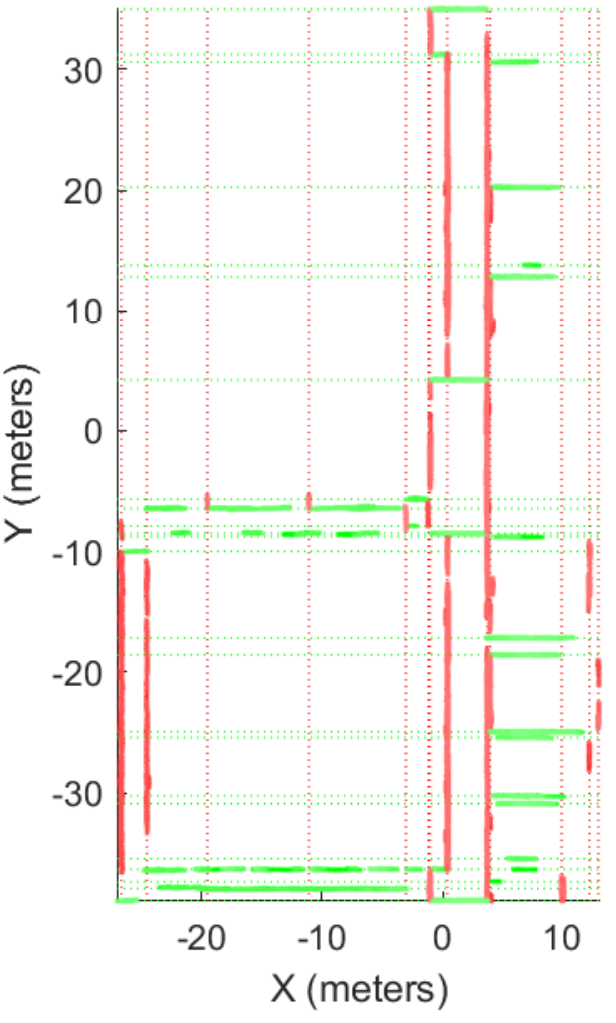}
        \caption{SL-LMS Input}
        \label{fig:sllmsout}
    \end{subfigure}
    \begin{subfigure}{0.3\textwidth}
        \includegraphics[width=0.9\textwidth]{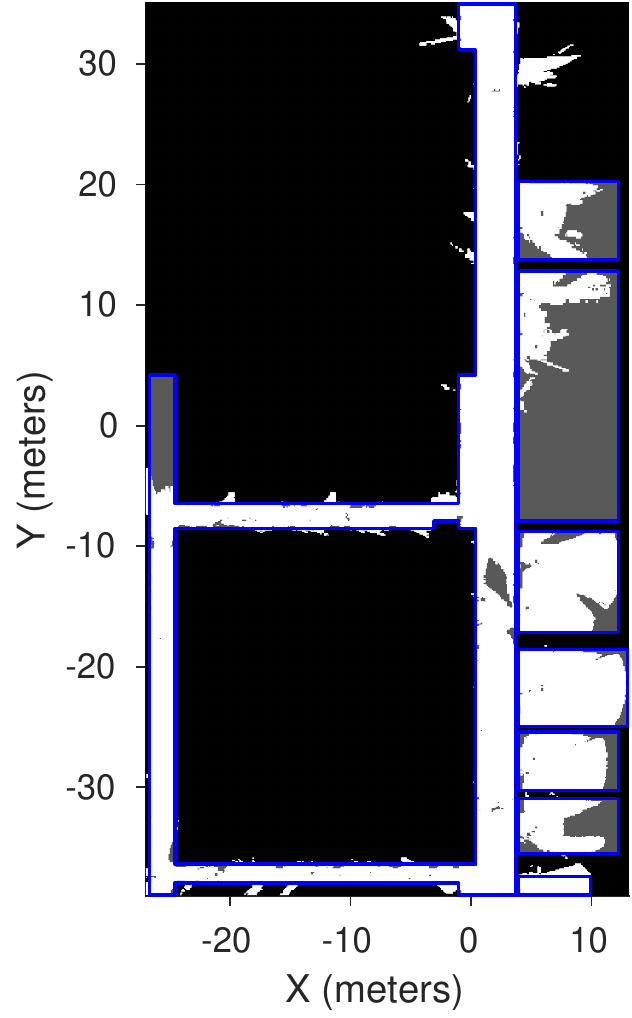}
        \caption{Free Space Speculation}
        \label{fig:overlay}
    \end{subfigure}
    \begin{subfigure}{0.3\textwidth}
        \includegraphics[width=0.9\textwidth,clip,trim={1.5in 0in 1.8in 0.3in}]{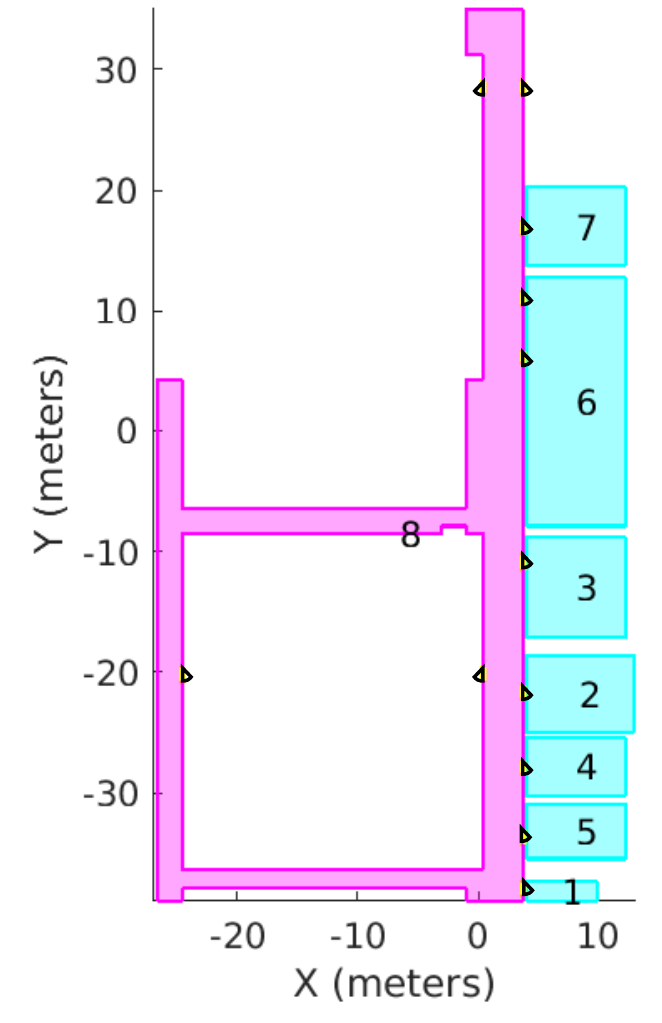}
        \caption{Semantic Floor Plan}
        \label{fig:semantic}
    \end{subfigure}
    \caption{A birds-eye perspective of the 3D reconstruction provided by \cite{shariati2018simultaneous} is shown in (a). Red and green dotted lines indicate the position of different layout planes perpendicular to the $x$ and $y$ axis, respectively. Each red and green point cloud illustrates the portion of its corresponding layout plane which is observed. A generated floor plan outlined in blue overlaid on top of the occupancy grid is given in (b). Known free cells are colored white while unobserved cells speculated to be free based on the floor plan are colored gray. Occupied cells and unobserved cells outside of the domain of the floor plan are colored black. The final semantically colored floor plan with labeled region is shown in (c). Cyan regions correspond to rooms, while magenta regions regions correspond to corridors. Open doorways on the borders of each region are indicated.}
    \label{fig:area3}
\end{figure*}

The method we present in this section builds upon the work of Shariati \etal \cite{shariati2018simultaneous}. In their paper, the authors present a SLAM approach that models partial observations of layout structures, including walls, floors, and ceilings, as segments residing on structural supporting \emph{layout planes}, which are modeled as axis-aligned surfaces with infinite extent. The output of their system is an optimized trajectory and the minimum set of layout planes necessary to explain all observed \emph{layout segments}. Each layout plane is parameterized by its position and orientation -- north, east, south, or west. This representation of indoor spaces leads to a simplified feature set for more robust localization and loop closure, and provides a strong prior for scene understanding. As is illustrated in Figure \ref{fig:sllmsout}, the primary shortcoming of such a representation, which we seek to address, is that although a human may be able to discern independent regions within this reconstruction, it is unclear how this may be done automatically since the true extent of individual walls and the meeting of perpendicular walls at corners is typically unobserved and ill defined. Furthermore, it is unclear how a robot could speculate about the unobserved free space in the environment given such a representation.

Although the approach described in the sequel can readily be extended to the multi-floor 3D case, for the sake of simplicity we describe the single-floor formulation of the problem, which will output a floor plan of the building. Note however that the system still analyzes a complete 3D point cloud and 3D trajectory to produce the distilled floor plan.

\subsection{Door Detection}

We first observe that in the context of buildings, doorways play a special role in scene understanding in that they signal a clear transition between two functionally disjoint spaces such as distinct rooms and hallways. This motivates us to develop a scheme to automatically detect these openings in walls.

\begin{figure*}
    \centering
    \includegraphics[width=\textwidth,clip,trim={0.7in 2.1in 0.5in 1.9in}]{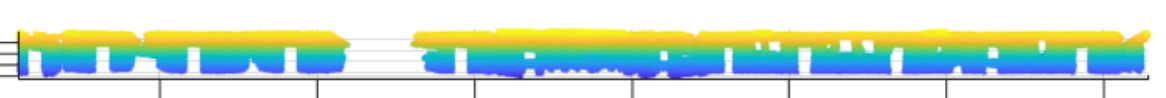}
    \includegraphics[width=0.965\textwidth,clip,trim={0.33in 0.12in 0in 0in}]{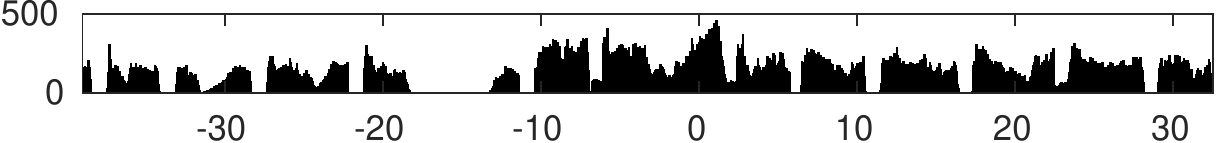}
    \caption{Result of merging the individual cloud segments associated with a particular layout plane (top). Histogram of projected points corresponding to the point cloud in Figure \ref{fig:segment} cropped at 2 meters (bottom). The distance between ticks along the axis is 10 meters. Histogram bin counts range from 0 - 500. }
    \label{fig:segment}
\end{figure*}

Given a layout plane we begin by accumulating all of the points associated with that plane as shown in 
Figure \ref{fig:segment}. Notice how the discontinuity of observations makes this problem of finding doors more challenging than simply looking for negative space. 

Each point set is cropped at 2 meters which is close to the typical door height and the remaining points are aggregated in a histogram as shown in Figure \ref{fig:segment}. We compute a smoothed gradient of the resulting signal and then convolve that result with a matched filter that is designed to detect 1 meter wide apertures. The existing system readily detects openings between 0.8 and 1.2 meters wide and can easily be extended to accommodate varying dimensions.
Note that the detected door openings are effectively children of the underlying layout planes. Once the layout has been determined these doorways help to define the functional transitions between different spaces.

\subsection{Layout Estimation}

We begin by recognizing that regions of indoor free space are typically enclosed by pairs of inward facing structures \ie north-facing wall to south-facing wall and east-facing wall to west-facing wall. Therefore, given a set of axis-aligned layout planes as well as their observed orientation, we enumerate all possible north-south and east-west pairs,
\begin{equation}
    \mathcal{P}_x = \mathcal{X}^+ \times \mathcal{X}^-
\end{equation}
\begin{equation}
    \mathcal{P}_y = \mathcal{Y}^+ \times \mathcal{Y}^-
\end{equation}
where $\mathcal{X}^+$ and $\mathcal{X}^-$ are defined as the set of east-facing plane positions and west-facing plane positions respectively. Similarly, $\mathcal{Y}^+$ and $\mathcal{Y}^-$ are defined in the same way for north and south-facing planes. Using these intermediary sets, we then enumerate all the possible rectangles that could be used to explain the free space
\begin{equation}
    \mathcal{R} = \mathcal{P}_x \times \mathcal{P}_y.
\end{equation}

This set however contains several different types of invalid rectangles including: those that have opposing faces which are outward facing; those whose length or width are too narrow (less than 1 meter for most indoor spaces); those which include portions of observed layout segments (projected to the ground plane after thresholding all points at a height greater than 2 meters) within their bounds; and those which include detected doorways within their bounds. Therefore, we prune the set of all such offending elements. It is interesting to note that this operation typically reduces the size of the original space of candidates by about $70$ - $95\%$, which greatly improves the speed of our algorithm.

This approach to defining rectangular regions is similar to the scheme employed by Xiao \etal \cite{Xiao2014} but here we leverage the fact that we are pairing structural planes with infinite extent rather than incomplete wall segments. More specifically if we consider the example environment shown in Figure \ref{fig:sllmsout} the system considers pairs of the infinite dotted lines shown rather than just the solid segments where direct evidence is available. This approach allows the scheme to effectively speculate in regions that have not been observed yet.


In addition to the planar reconstruction, \cite{shariati2018simultaneous} also provides a voxel map reconstruction of the environment, from which we can sample at a particular height in order to determine which cells each rectangle in $\mathcal{R}$ spans. Each voxel is $0.1$ meters on side.

At this point, we observe that the problem we are presented with can be phrased as a set cover problem. We are given a universe $F_h$ of $n$ free space cells in the 2D occupancy grid -- generated by sampling from the 3D voxel map at a height $h$ -- each covered by at least one $R \in \mathcal{R}$, and a list of $R_1, \dots, R_m \in \mathcal{R}$ rectangle subsets of $F_h$, each with its own weight defined as the total number of free, occupied, and unobserved cells in the grid it covers. What we would like is to select the a collection of rectangles, $\mathcal{C}$, of minimum total weight, whose union is equal to all of $F_h$. Minimizing this objective should, in principle, select those candidates which explain as much of the free space as they can, while also yielding the simplest explanation of the space. The set cover problem is, of course, NP-Complete, however effective greedy solutions have been developed and we exploit one of these.

We construct the cover $\mathcal{C}$ of rectangles, by iteratively making the following greedy choice: select the rectangle $R_i$ that minimizes
\begin{equation}
    \frac{A_i}{| R_i \cap D |}
\end{equation}
until no free space voxels remain uncovered, where $A_i$ denotes the sum of the number of free, occupied, and unobserved voxels within the span of $R_i$, and $D$ is the set of remaining uncovered free voxels. If there happen to be two or more rectangles with the same ratio, we choose the one with the largest $A_i$. This algorithm has the interesting property that the cover selected has weight within a factor $O(\log d^*)$ of the optimal, where $d^* = \max_i | S_i |$ \cite{kleinberg2006algorithm}.

Given this cover, a floor plan can be generated by computing the union of all rectangles in $\mathcal{C}$. An example of such a floor plan can be seen in Figure \ref{fig:overlay}. Notice that our segment and doorway collision constraint on $\mathcal{R}$ results in the generation of functionally disjoint regions. These regions may also be given unique identifiers as illustrated in Figure \ref{fig:semantic}. It is important to mention that these regions are subject to one filtering criteria, which is that no region my have a ratio of total cells spanned to free cells spanned greater than 1000. This threshold may be tuned in order to limit the desired degree of risk in the speculation.

Based on which layout planes form the faces of each region, we can also reason as to which doorways act as transitions between pairs of adjacent regions given each region's well defined boundary. These doorways are highlighted in the semantically annotated version of the floor plan shown in Figure \ref{fig:semantic}.

Note that this optimization procedure seeks to find the simplest set of boxes that explains the available data which encourages the system to expand corridors and rooms since this allows it to explain larger regions with fewer primitives. In contrast the optimization in  \cite{Xiao2014} was designed for situations where the space was completely scanned so the optimization penalizes primitives that include unexplored voxels.

\subsection{Semantic Labeling}

For any functional interpretation of space, it is important to understand what each region represents. In our classification scheme, we distinguish between two types of spaces: rooms and corridors. While this label space may not be comprehensive, we argue that it does capture the general purpose of most types of space -- either the space itself acts as a transition, or the space is itself a terminal point where some particular event or action takes place. Observe however, that these categories, rooms in particular, can each be readily extended to include subcategories such as office, kitchen, etc.

The layout estimation algorithm described in the previous section produces a floor plan $\mathcal{L}$ comprised of $k$ disjoint regions parameterized by sets of vertices. For each of the $k$ regions we can compute several features to describe the particular space, including the area, perimeter and aspect ratio. Recognizing that the outer boundary of most rooms are typically close to square, the feature which yields the largest information gain between the two classes is the turning distance \cite{arkin1991efficiently} between the region's outer boundary, with its perimeter normalized to 1, and the unit square. This quantity turns out to be quite useful as it implicitly captures the magnitude of various other attributes at once such as the number of sides and the aspect ratio. 

Using these features, it is possible to use the following classifier to discriminate between the two types of regions:
\begin{equation}
    h(x) = \begin{cases}
        \text{if } \texttt{perim}(x) < 60 \text{ and } & \textbf{room} \\ \; \texttt{turndist}(x,\texttt{square}) < 1, & \\ 
        \text{else} & \textbf{corridor}
    \end{cases}
\end{equation}

An example of a semantically labeled floor plan can be seen in Figure \ref{fig:semantic}. While this hand crafted classifier is quite simple, it does represent a useful baseline against which other, more sophisticated, schemes could be compared.

We feel it is important to emphasize that in this approach we are \emph{not} performing semantic segmentation, but rather a semantic labeling of high-level regions in a water-tight model. While it may help provide a description of individual observations, semantic segmentation of a point cloud or an occupancy grid does not produce any abstraction or model that may be useful for higher-level reasoning.

\section{Experimental Results}

\begin{figure*}[t]
    \centering
    \begin{subfigure}{0.4\textwidth}
        \centering
        \begin{subfigure}{\textwidth}
            \centering
            \includegraphics[width=0.45\textwidth]{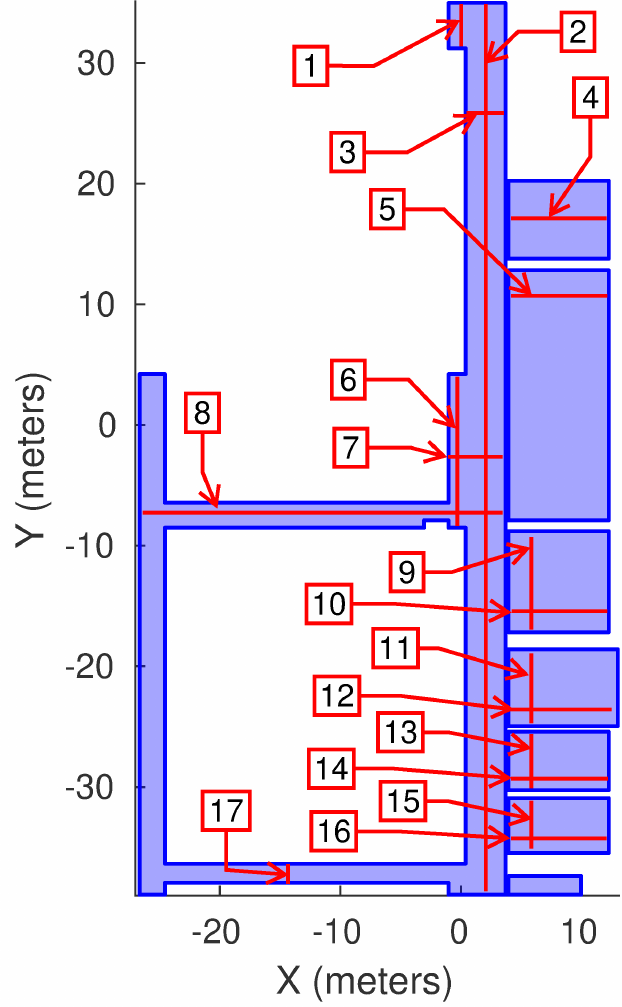}
            \caption{Floor Plan A}
        \end{subfigure}
        \begin{subfigure}{\textwidth}
            \includegraphics[width=0.85\textwidth]{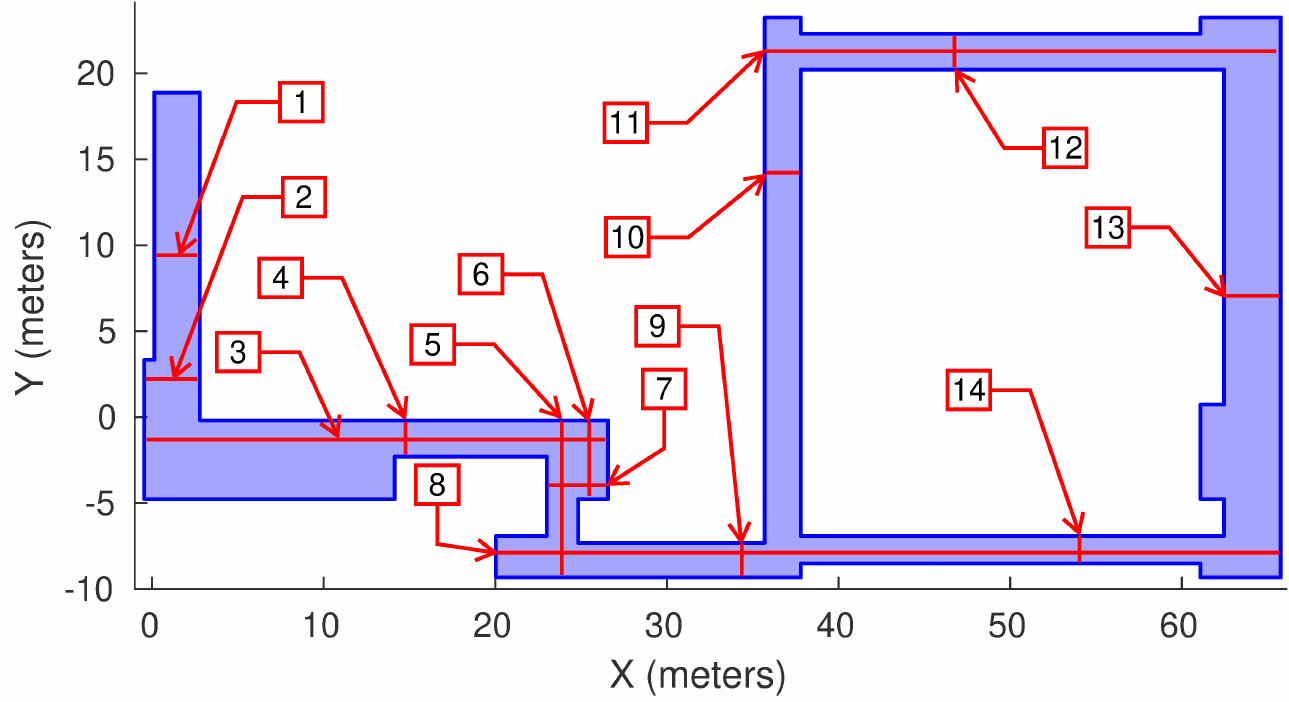}
            \caption{Floor Plan B}
        \end{subfigure}
    \end{subfigure}
    \begin{subfigure}{0.54\textwidth}
        \begin{tabular}{|c|c|c|c|c|c|c|}
        \hline
        \multirow{2}{*}{ID} & \multicolumn{3}{c|}{Floor Plan A} & \multicolumn{3}{c|}{Floor Plan B} \\
        & GT & FP & $\Delta$ & GT & FP & $\Delta$ \\ \hline
        1 & 3.62 & 3.75 & 0.13 & 2.72 & 2.66 & 0.06 \\ \hline
        2 & 73.15 & 73.91 & 0.76 & 3.23 & 3.25 & 0.02 \\ \hline
        3 & 3.34 & 3.32 & 0.02 & 27.13 & 27.04 & 0.09 \\ \hline
        4 & 9.03 & 8.31 & 0.72 & 2.17 & 2.10 & 0.07 \\ \hline
        5 & 8.46 & 8.31 & 0.15 & 9.15 & 9.14 & 0.01 \\ \hline
        6 & 12.60 & 12.73 & 0.13 & 4.61 & 4.57 & 0.04 \\ \hline
        7 & 4.72 & 4.72 & 0 & 3.59 & 3.58 &  0.01 \\ \hline
        8 & 29.99 & 30.34 & 0.35 & 45.04 & 45.72 & 0.68 \\ \hline
        9 & 8.41 & 8.38 & 0.04 & 2.02 & 2.02 & 0 \\ \hline
        10 & 8.48 & 8.31 & 0.17 & 2.10 & 2.10 & 0 \\ \hline
        11 & 6.33 & 6.37 & 0.04 & 30.01 & 30.06 & 0.05 \\ \hline
        12 & 9.02 & 9.06 & 0.04 & 2.00 & 2.08 & 0.08 \\ \hline
        13 & 4.80 & 4.84 & 0.04 & 3.30 & 3.31 & 0.01 \\ \hline
        14 & 8.44 & 8.31 & 0.13 & 1.68 & 1.6 & 0.08 \\ \hline
        15 & 4.56 & 4.56 & 0 & && \\ \hline
        16 & 8.57 & 8.31 & 0.26 & && \\ \hline
        17 & 1.47 & 1.58 & 0.11 & && \\ \hline \hline
        Mean Err. & \multicolumn{3}{c|}{0.18} & \multicolumn{3}{c|}{0.09} \\ \hline \hline
        Mean \% Err. & \multicolumn{3}{c|}{1.98} & \multicolumn{3}{c|}{1.31} \\ \hline
        \end{tabular}
    \end{subfigure}
    \caption{Floor plans A and B (left) annotated with locations of ground truth measurements. Differences between ground truth measurements and floor plan estimates are provided in table (right). All values are given in meters. }
    \label{fig:accuracy}
\end{figure*}

\begin{figure*}[p]
\begin{tabular}{ c | >{\centering\arraybackslash}m{1.7in} >{\centering\arraybackslash}m{1.7in} >{\centering\arraybackslash}m{1.7in} }
     Area ID & SL-LMS Input & Free Space Speculation & Semantic Floor Plan \\ \hline
     \rule{0pt}{0.8in} Area 1 & \includegraphics[height=1.7in,width=1.7in,keepaspectratio,clip,trim={1.5in 0in 1.7in 0.3in}]{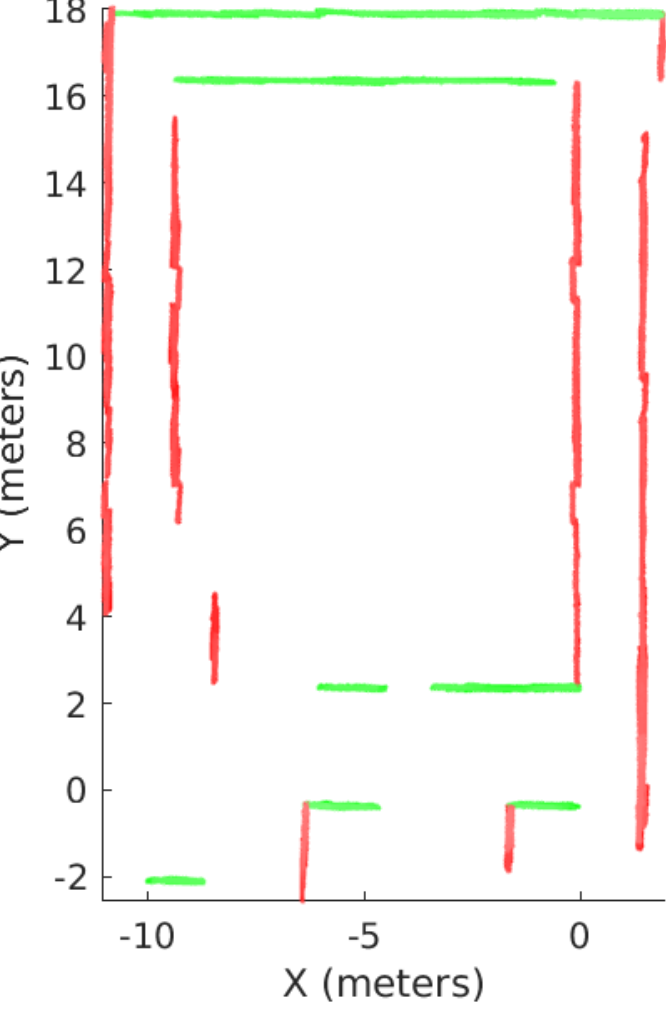} & \includegraphics[height=1.7in,width=1.7in,keepaspectratio]{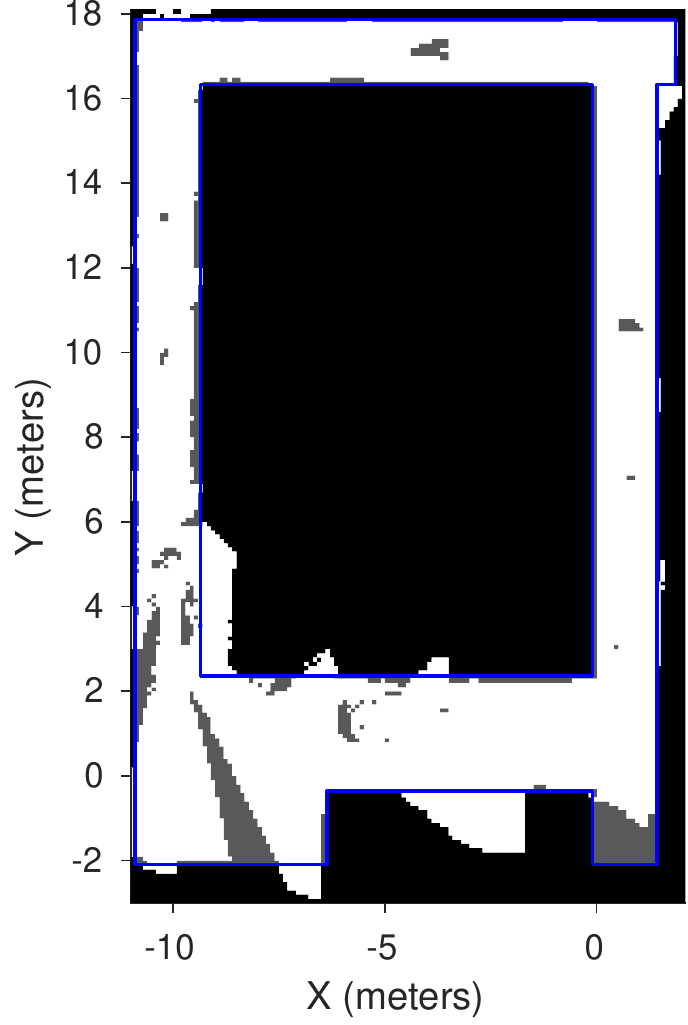} & \includegraphics[height=1.7in,width=1.7in,keepaspectratio]{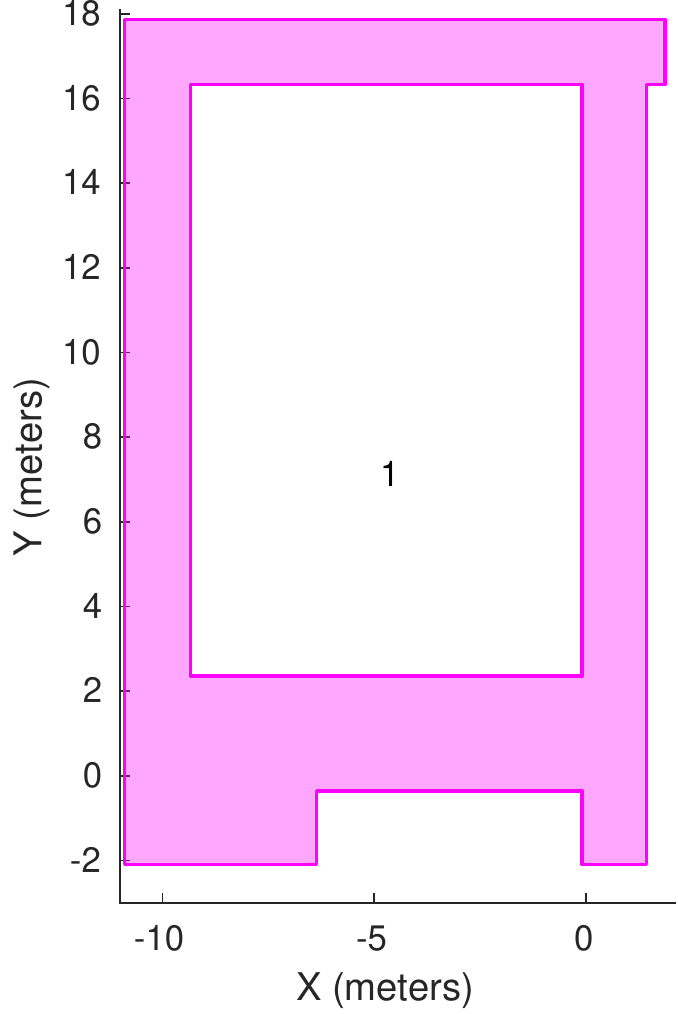} \\
     Area 2 & \includegraphics[height=1.7in,width=1.7in,keepaspectratio,clip,trim={0.3in 0in 0.5in 0in}]{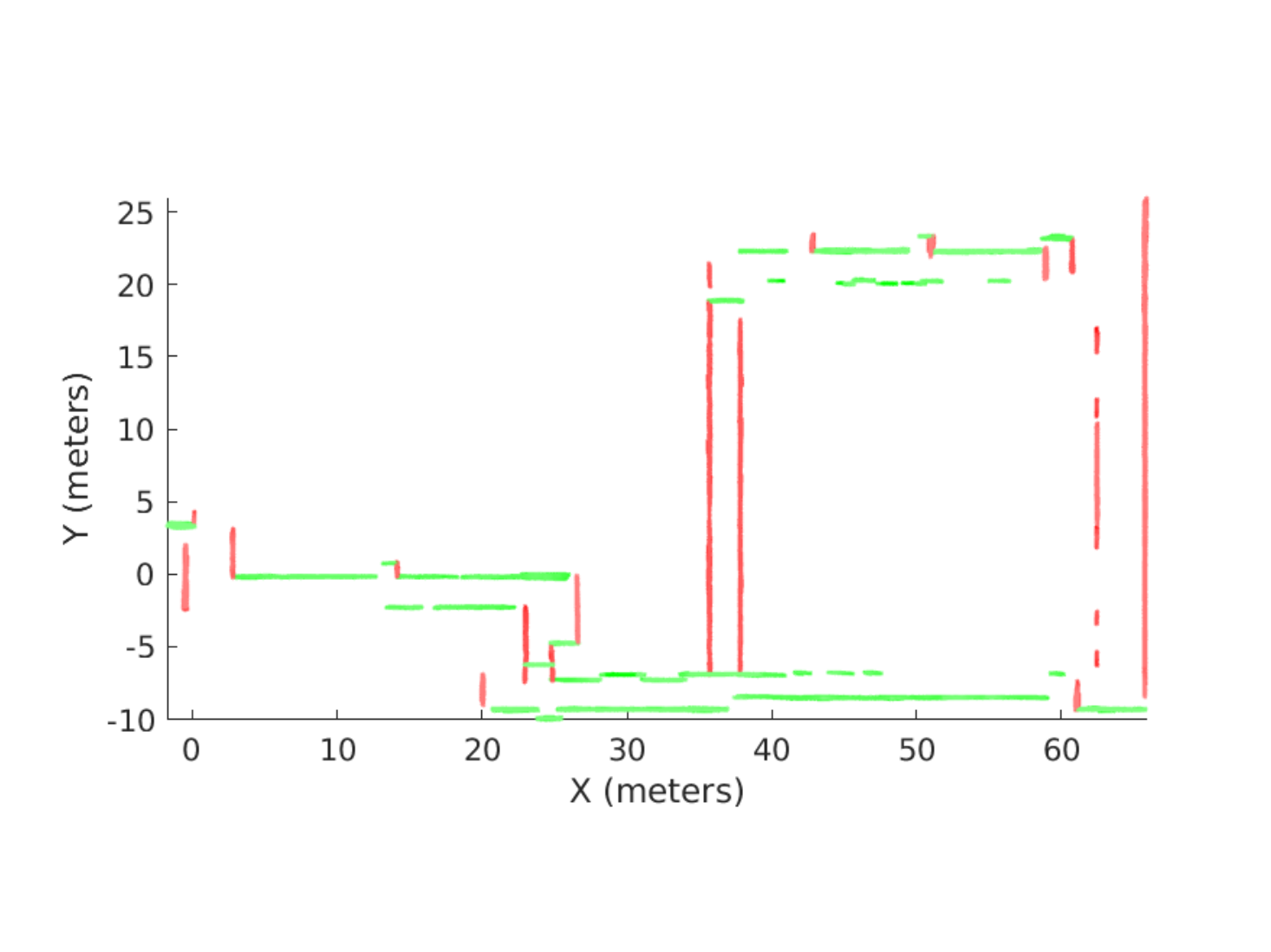} & \includegraphics[height=1.7in,width=1.7in,keepaspectratio]{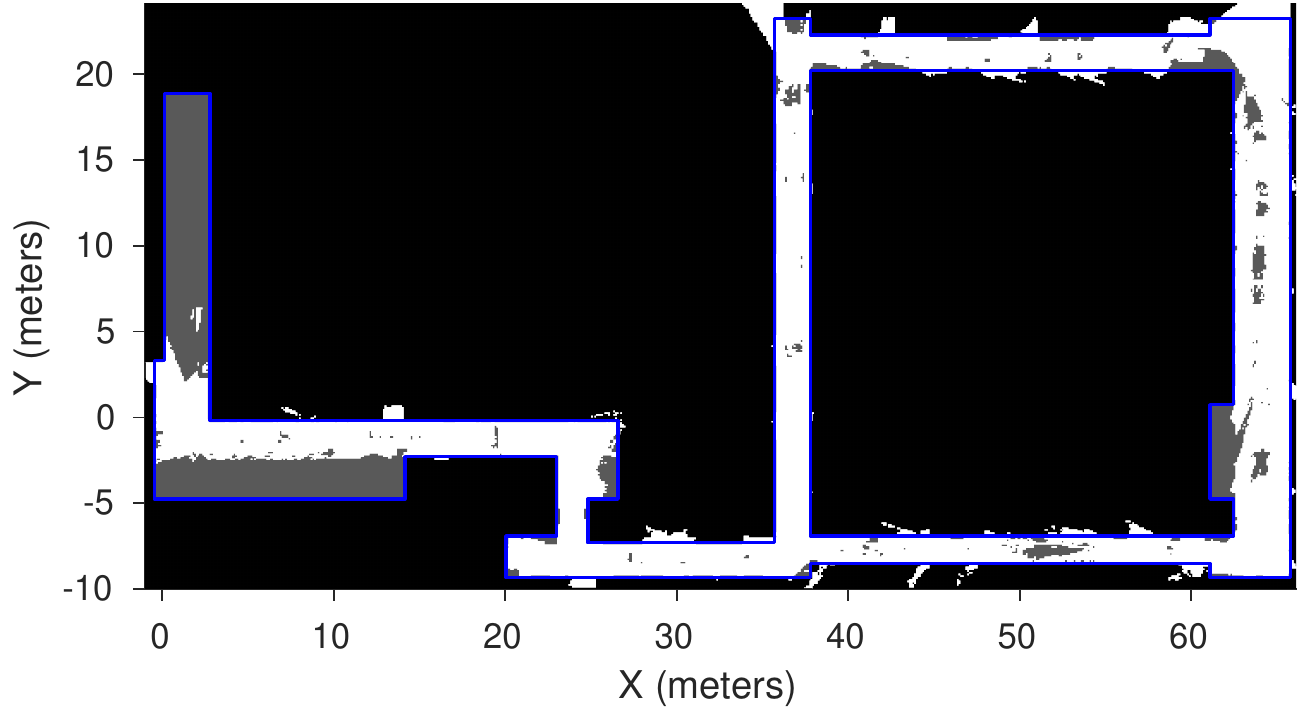} & \includegraphics[height=1.7in,width=1.7in,keepaspectratio,clip,trim={0.25in 0.7in 0.5in 1in}]{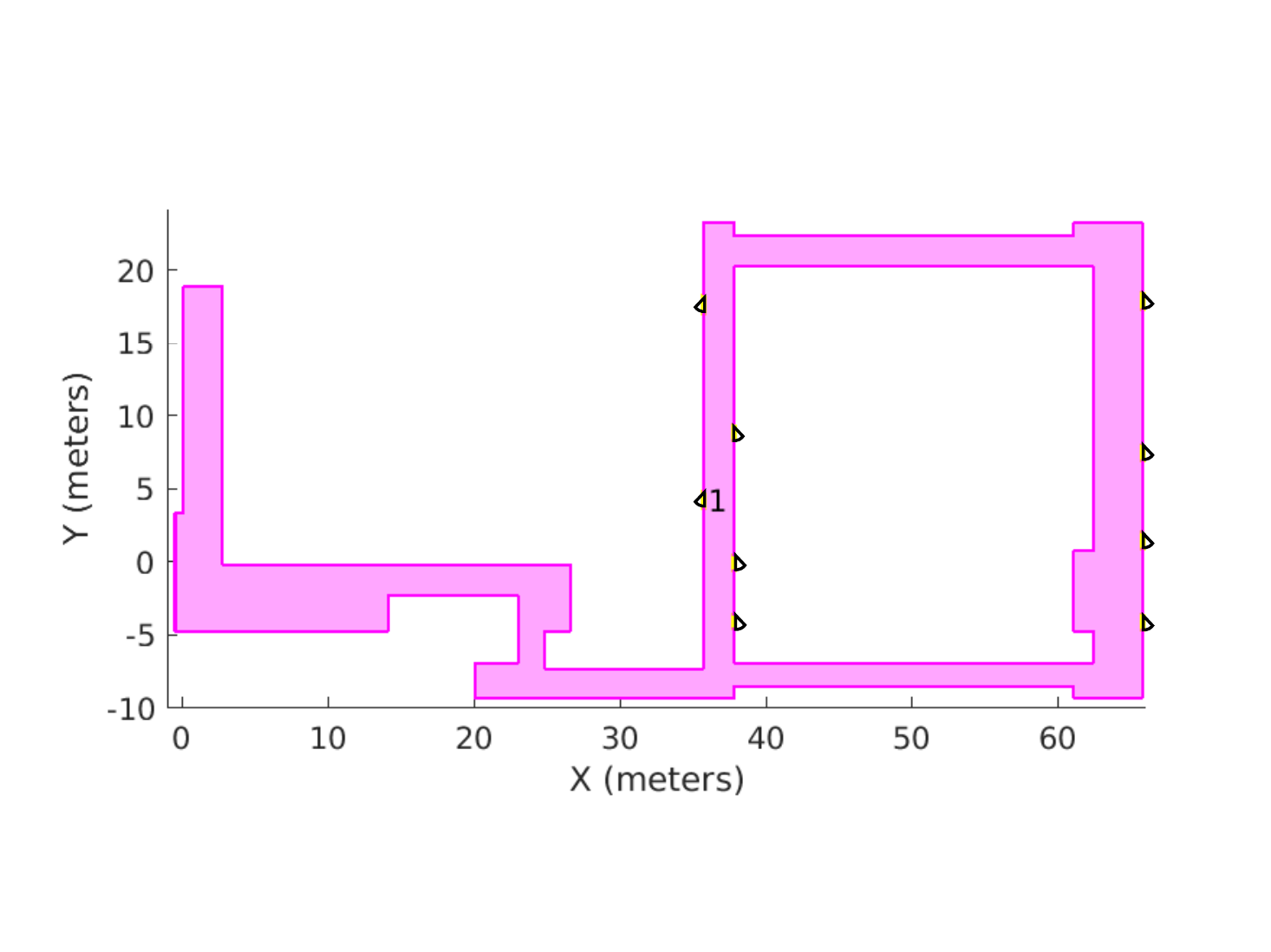} \\ 
     Area 3 & See Figure \ref{fig:sllmsout} & See Figure \ref{fig:overlay} & See Figure \ref{fig:semantic} \\
     Area 4 & \includegraphics[height=1.7in,width=1.7in,keepaspectratio,clip,trim={0.75in 0in 1.2in 0.25in}]{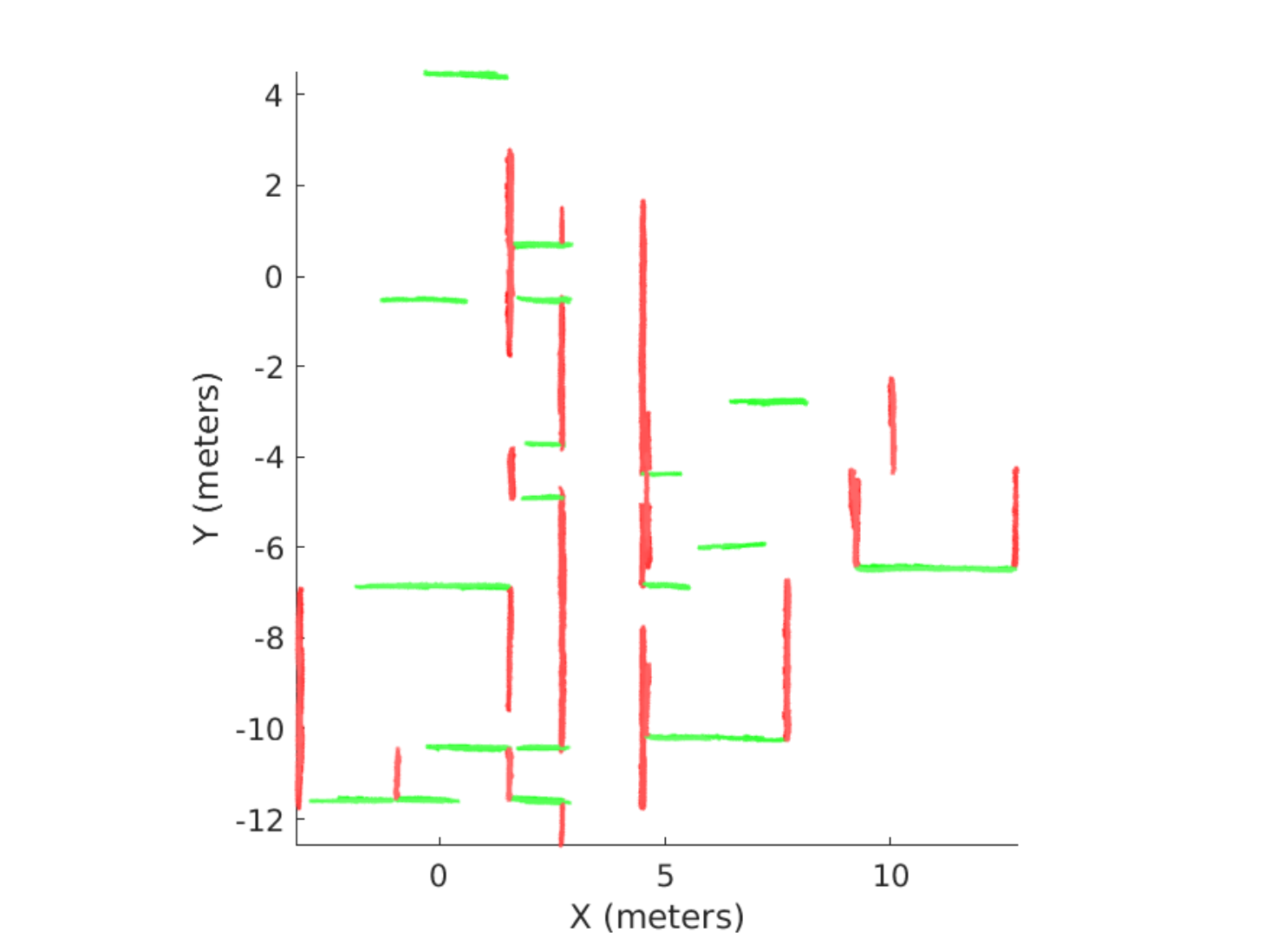} & \includegraphics[height=1.7in,width=1.7in,keepaspectratio]{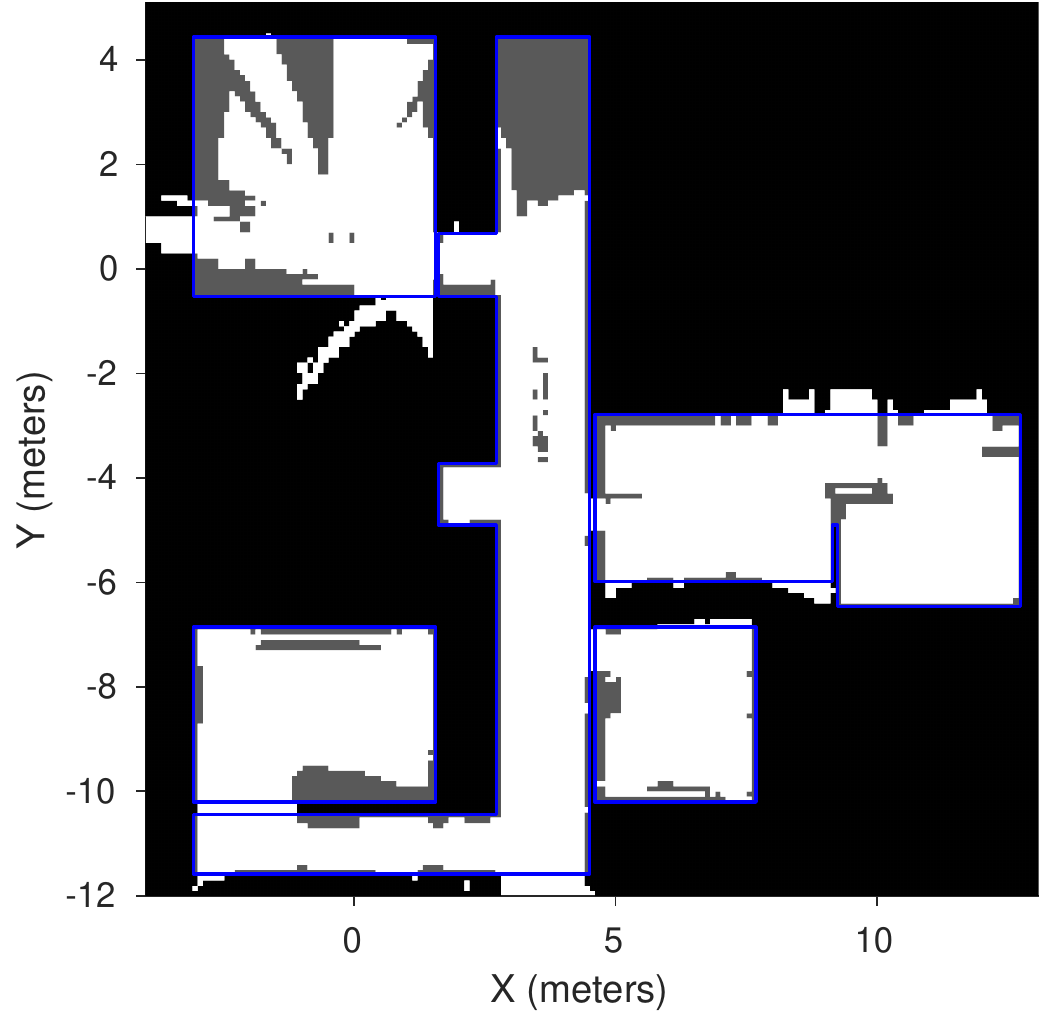} & \includegraphics[height=1.7in,width=1.7in,keepaspectratio,clip,trim={0.8in 0in 1.2in 0.4in}]{figures/results/fla-blackbox_2018-05-23-19-05-15_3rooms_explore/semantic_floorplan.pdf} \\
     Area 5 & \includegraphics[height=1.7in,width=1.7in,keepaspectratio,clip,trim={2.4in 0in 2.4in 0.25in}]{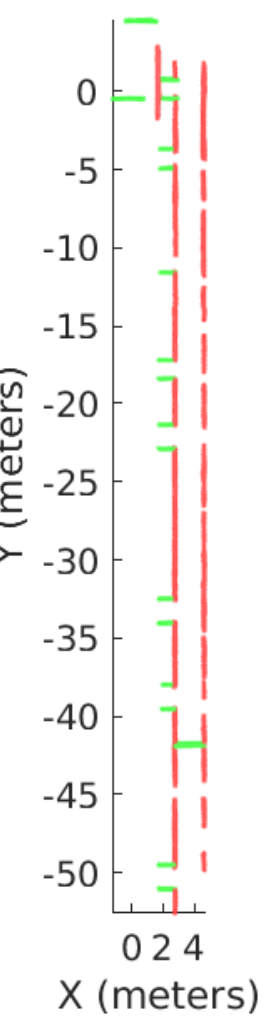} & \includegraphics[height=1.7in,width=1.7in,keepaspectratio]{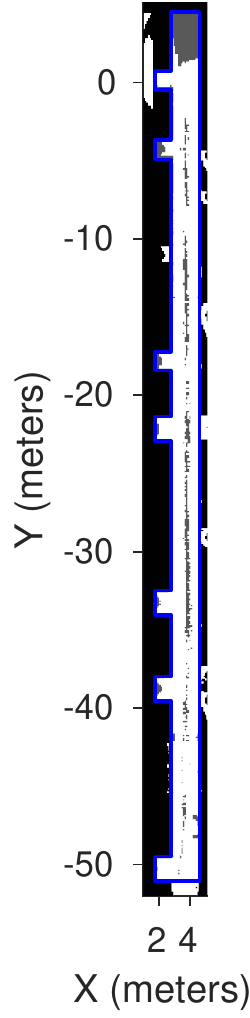} & \includegraphics[height=1.7in,width=1.7in,keepaspectratio]{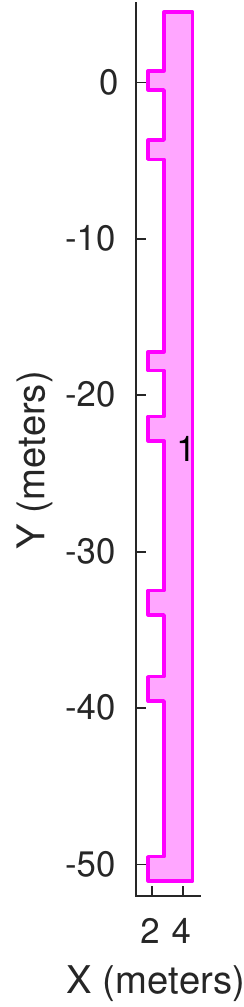} \\
     Area 6 & \includegraphics[height=1.7in,width=1.7in,keepaspectratio,clip,trim={0.4in 1.1in 0.5in 1.7in}]{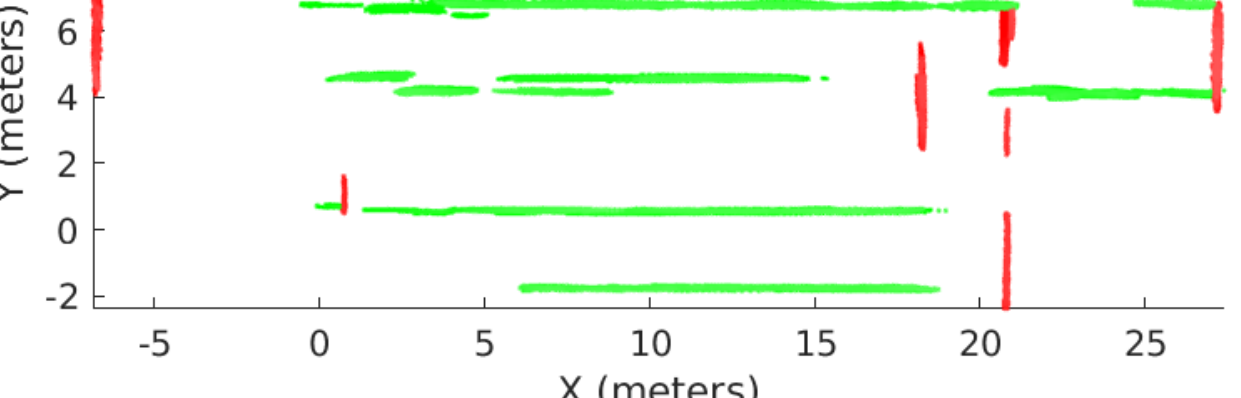} & \includegraphics[height=1.7in,width=1.7in,keepaspectratio]{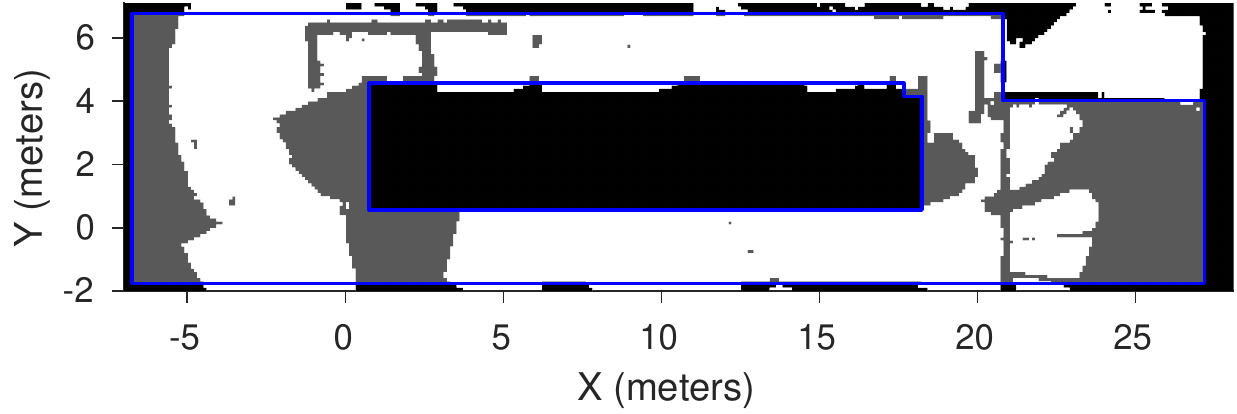} & \includegraphics[height=1.7in,width=1.7in,keepaspectratio]{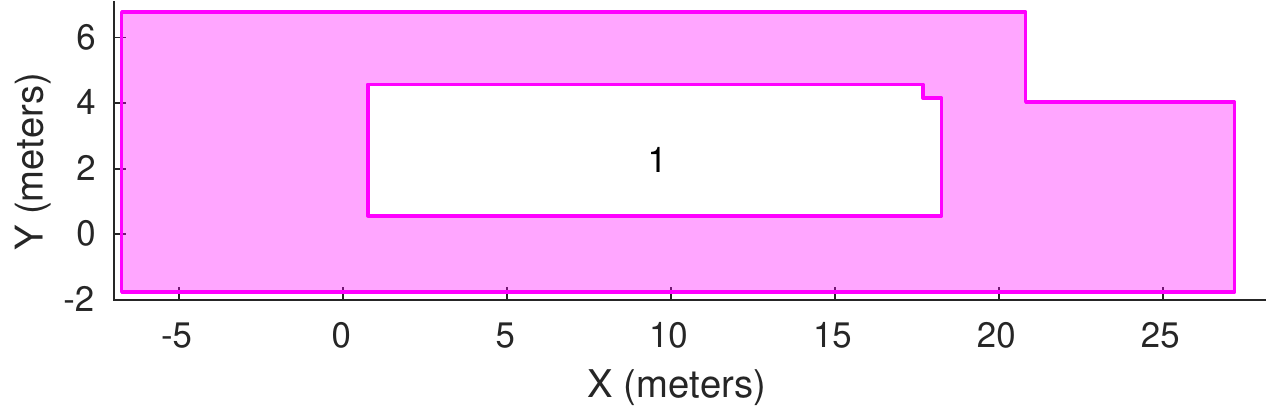} \\ 
\end{tabular}
\caption{Batch floor plan generation results in several indoor environments. The first column shows the input provided by \cite{shariati2018simultaneous}. The second column illustrates the difference between occupied, free, and speculated free space. The third column shows the semantic floor plan with doors, corridors, and rooms highlighted in yellow, magenta, and cyan respectively. }
\label{fig:composite}
\end{figure*}

In order to evaluate the efficacy of the proposed scheme we applied it to a number of extended indoor environments. Our data sets consisted of range images, stereo images and inertial measurements acquired with the sensor system shown on the robot in Figure \ref{fig:falcon}. Figure \ref{fig:area3} and Figure \ref{fig:composite} show the results of applying the interpretation in a batch mode on various data sets. In each of these cases the system was able to correctly infer the large scale building layout and partition the space into rooms and corridors. It was also able to correctly detect doorways which are indicated on the figures. These figures also compare the inferred free space area with the free space area that is actually observed to provide an indication of the systems ability to speculate about unexplored regions.

The results show the system's ability to infer the presence of structure which is not directly observed. For instance, in Area 3, the algorithm is able to use the easternmost plane of Rooms 3, 4, and 5, in order to infer the presence of a back wall in Rooms 6 and 7, and cover the free space observed in them. Also notice that the use of these rectangular primitives allows the system to approximate more complicated structure such as that seen in Room 4 of Area 4, which would have otherwise been lost in direct geometric model fitting schemes that would seek to approximate the entire space with a single cuboid.

In order to provide a more quantitative evaluation of the scheme we took measurements of the dimensions of a subset of the rooms and corridors used in our experiments using a hand held laser range finder and compared these measurements with the dimensions predicted by our automated layout scheme. These results are presented in Figure \ref{fig:accuracy}. The results indicate that over all the dimensions that were considered the measurements and the predictions agreed on average within 2\%.


In a second set of experiments we run the scene interpretation scheme in an online manner at regular intervals ($10$ - $20$ seconds) in the data set to provide an understanding of how a robots' concept of the space would evolve as it moved through the environment.

\begin{figure*}
    \centering
    \begin{subfigure}{\textwidth}
        \centering
        \includegraphics[width=0.25\linewidth]{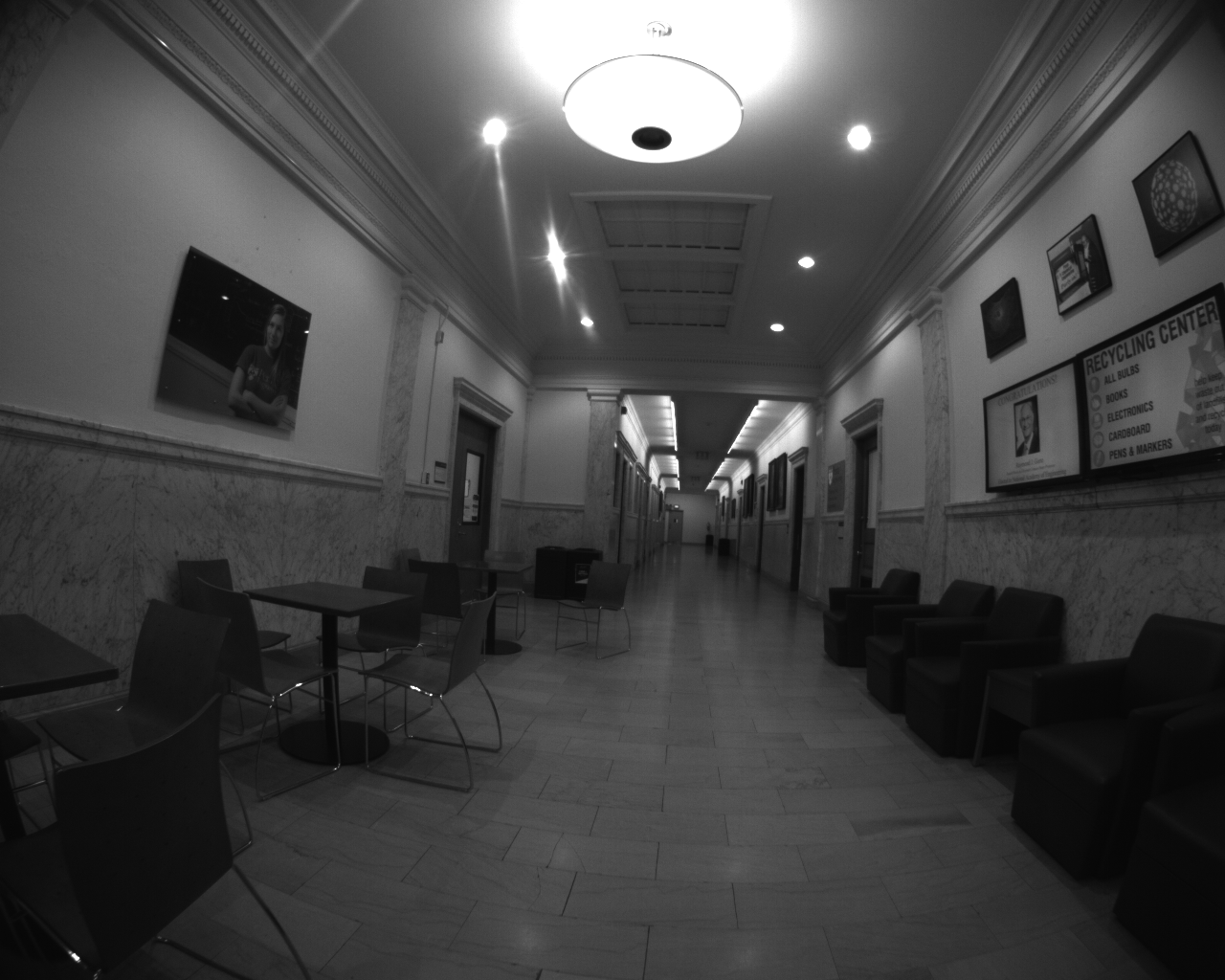}\includegraphics[width=0.25\linewidth]{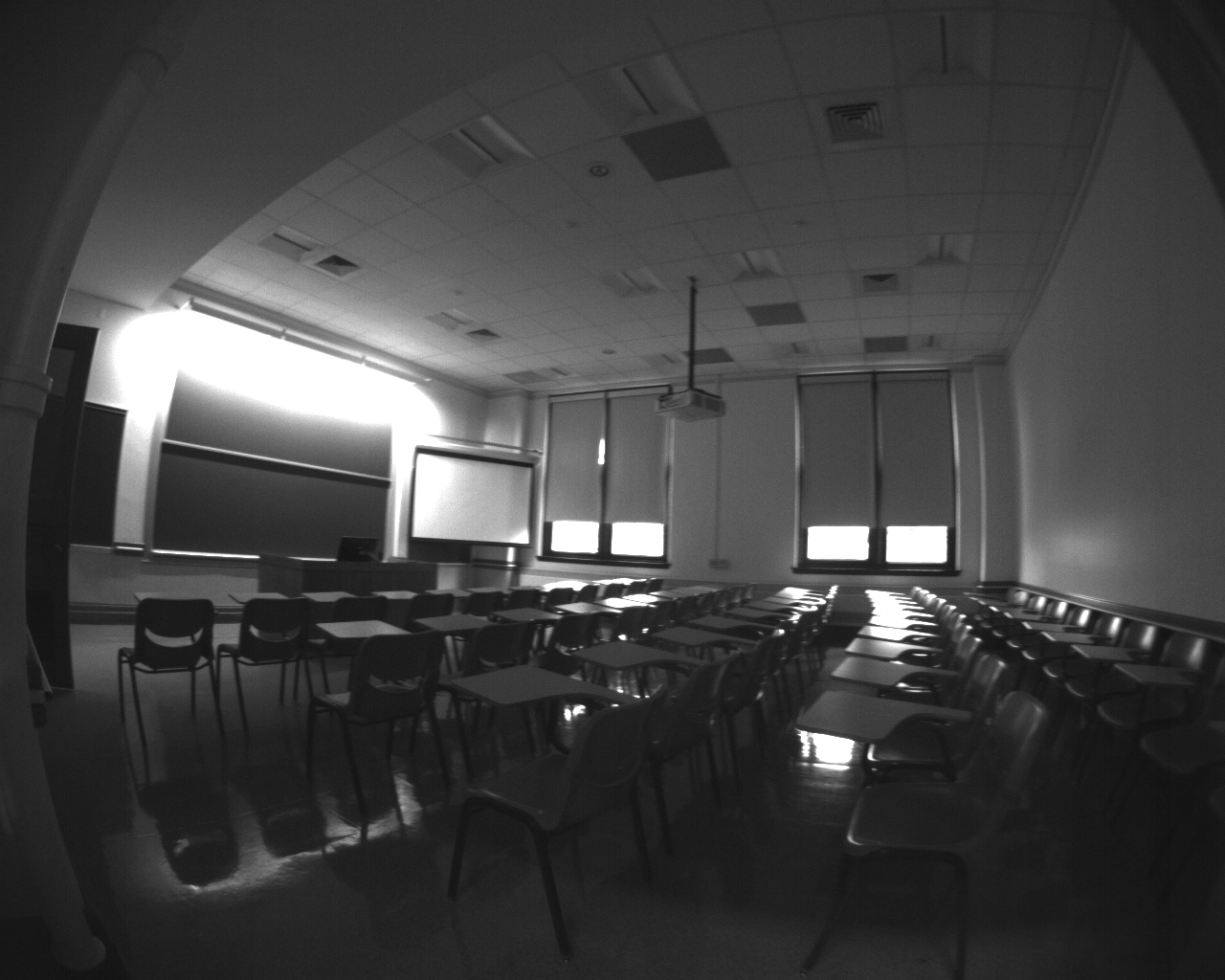}\includegraphics[width=0.25\linewidth]{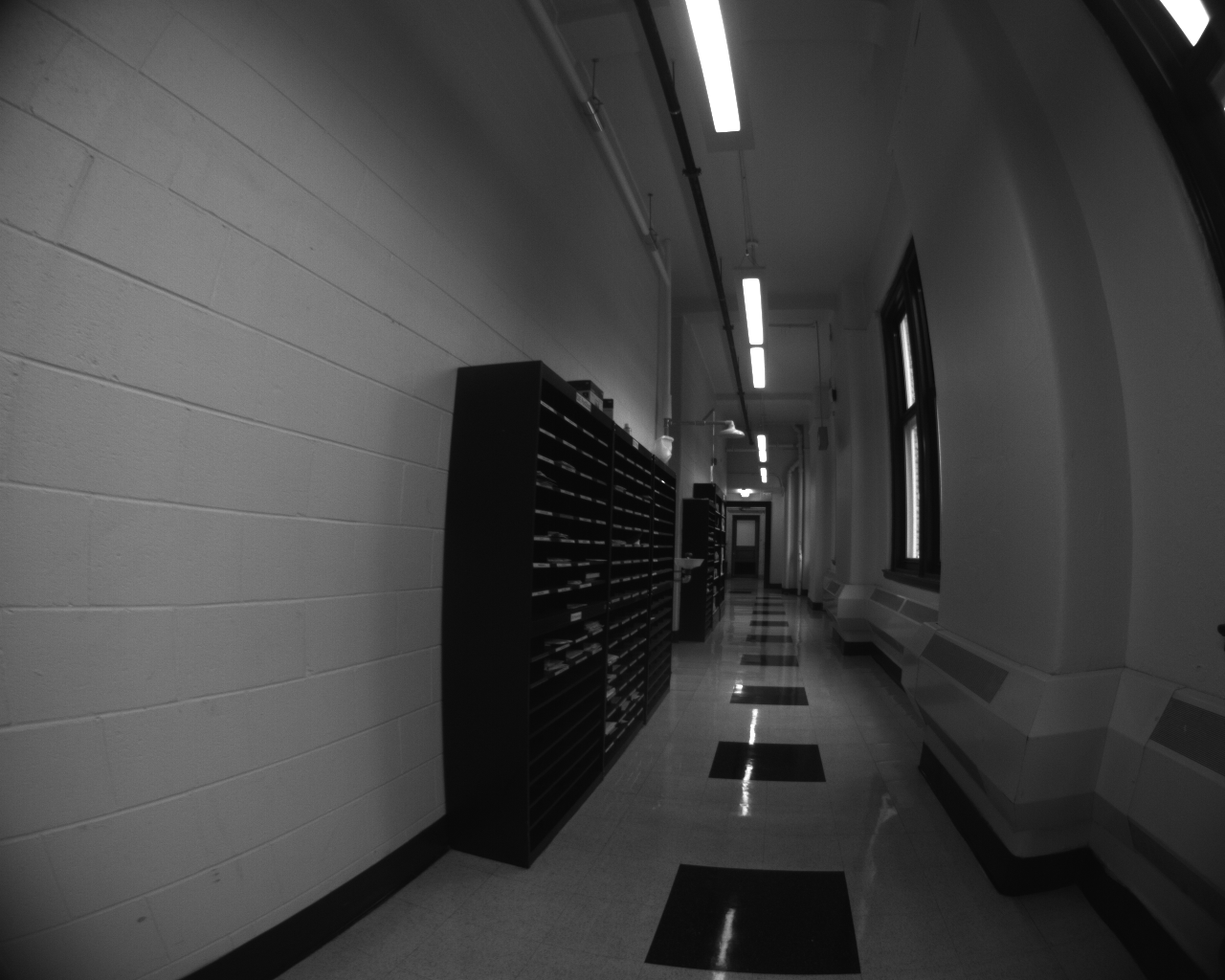}\includegraphics[width=0.25\linewidth]{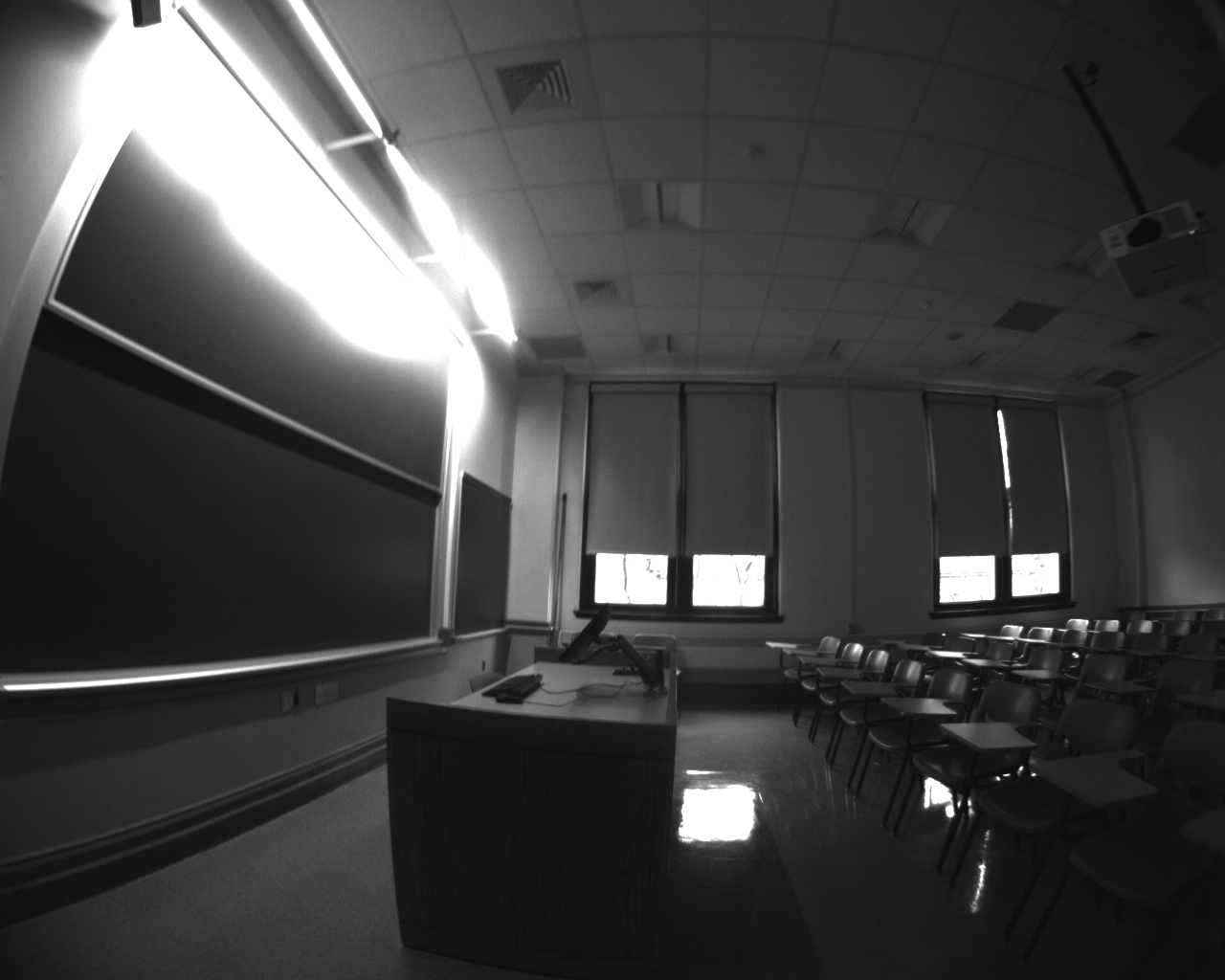}
        \caption{Sample Images}
        \label{fig:online1images}
    \end{subfigure}
    \begin{subfigure}{\textwidth}
        \centering
        \includegraphics[width=0.24\linewidth,height=2in,keepaspectratio,clip,trim={0.8in 0in 1.2in 0.25in}]{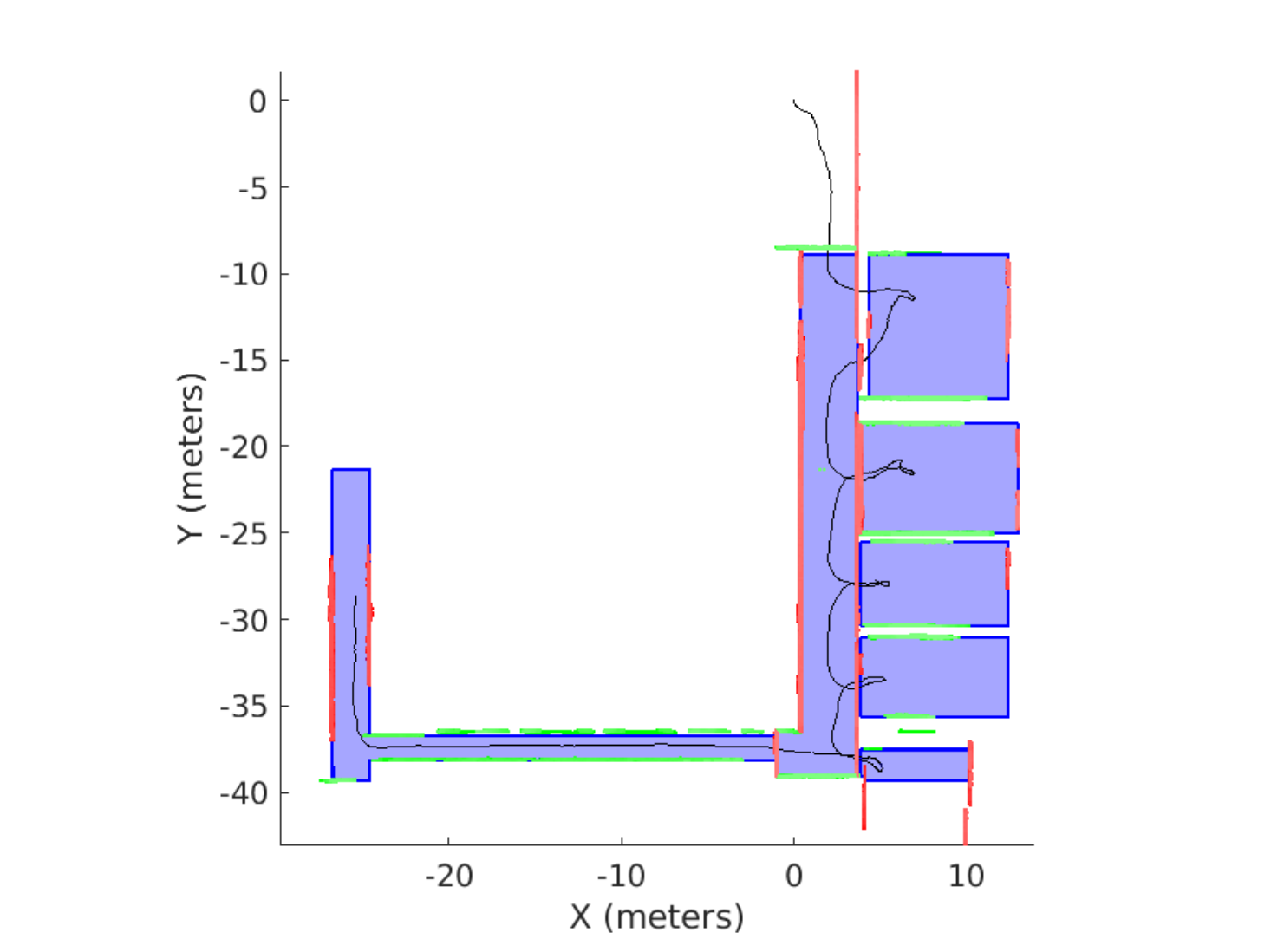}\quad\includegraphics[width=2in,height=2in,keepaspectratio,clip,trim={1.1in 0in 1.3in 0.25in}]{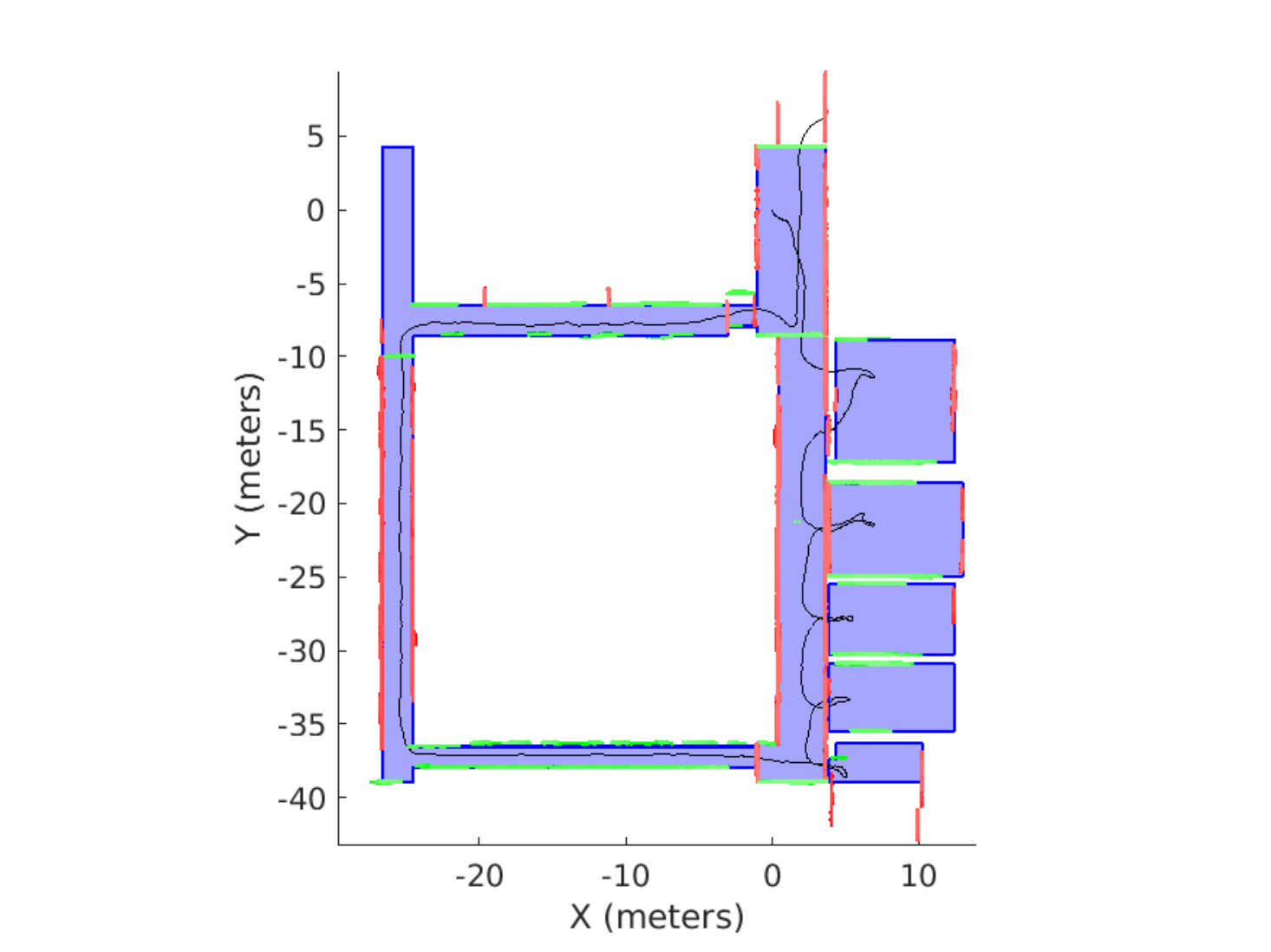}\quad\includegraphics[width=0.24\linewidth,height=2in,keepaspectratio,clip,trim={1.4in 0in 1.75in 0.5in}]{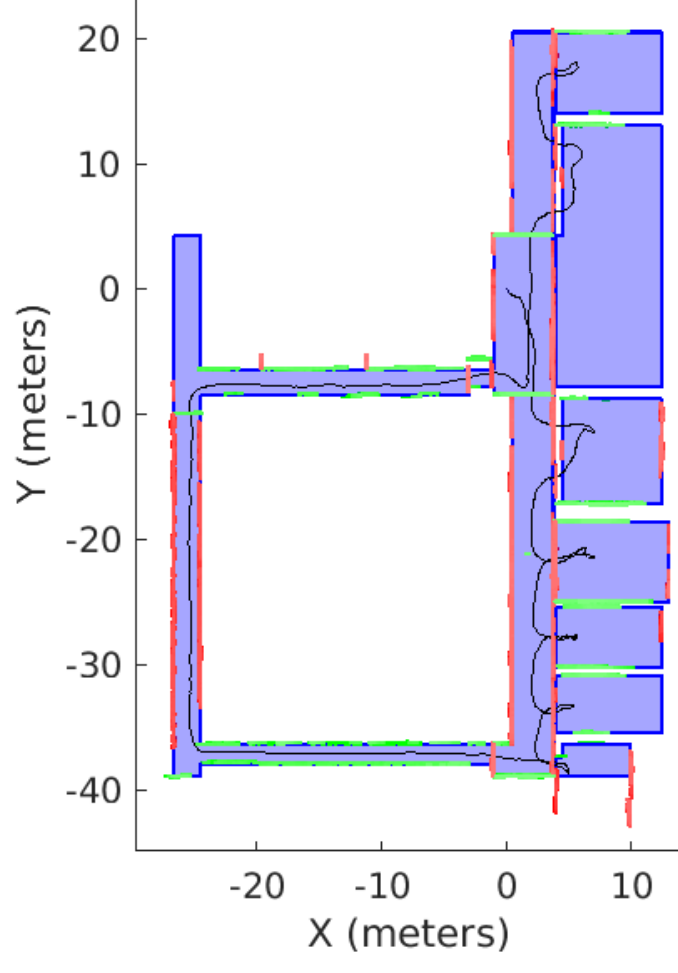}\quad\includegraphics[width=0.24\linewidth,height=2in,keepaspectratio,clip,trim={1.7in 0in 2in 0.5in}]{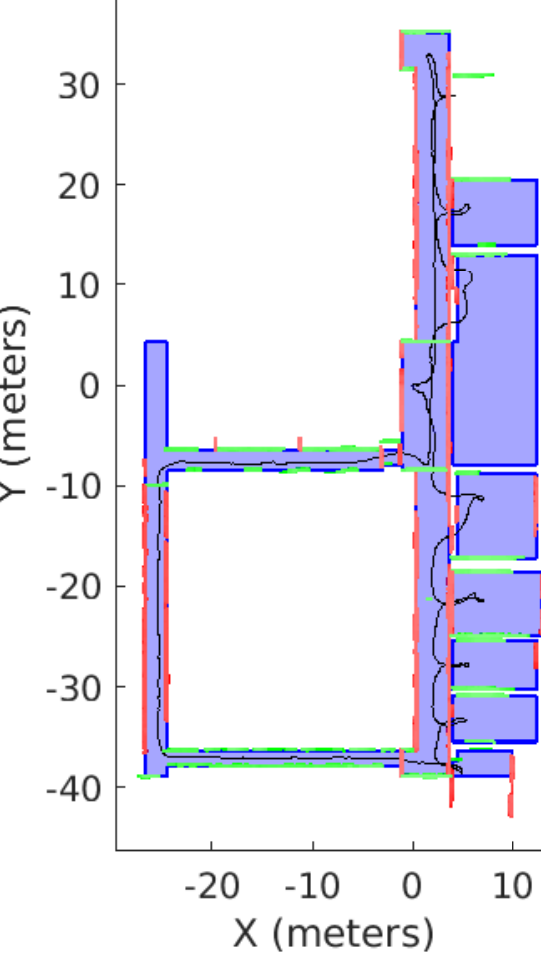}
        \caption{Estimated floor plan at $t=1000$, $1400$, $1600$, and $2075$ seconds.}
        \label{fig:online1}
    \end{subfigure}
    \caption{Online estimation results for Building A. Total distance traveled $=247$ meters. }
    \label{fig:my_label1}
\end{figure*}

\begin{figure*}
    \centering
    \begin{subfigure}{\textwidth}
        \centering
        \includegraphics[width=0.25\linewidth]{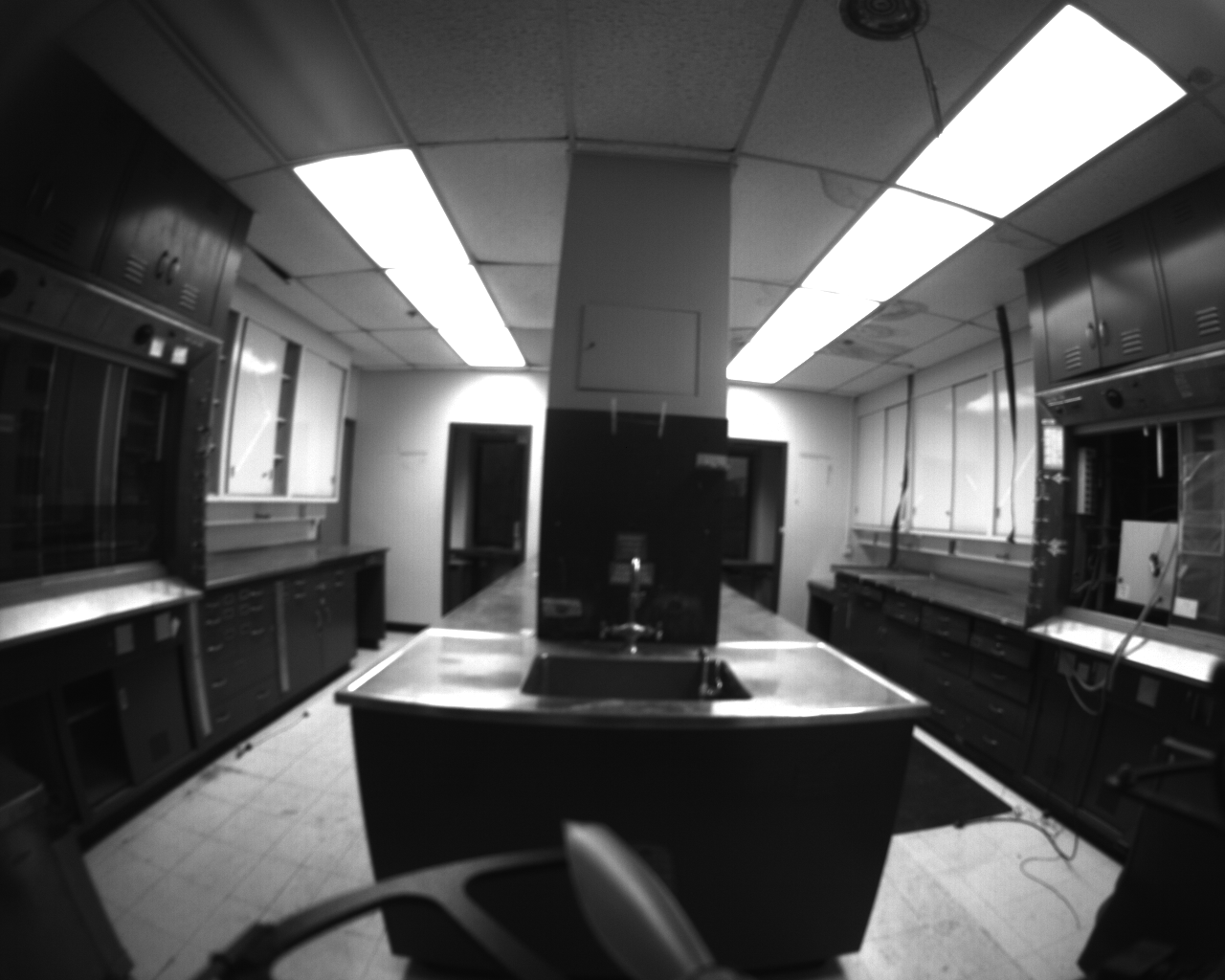}\includegraphics[width=0.25\linewidth]{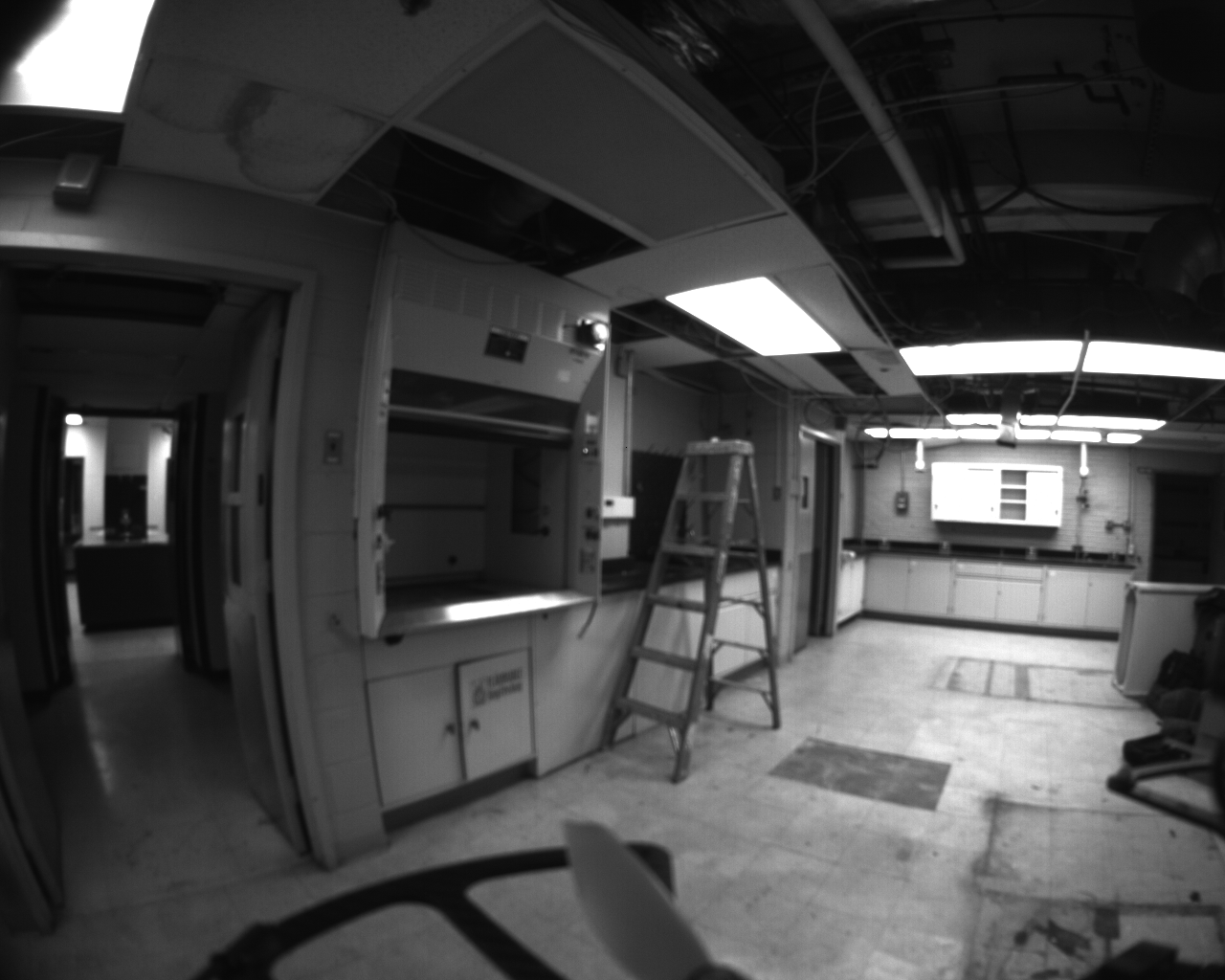}\includegraphics[width=0.25\linewidth]{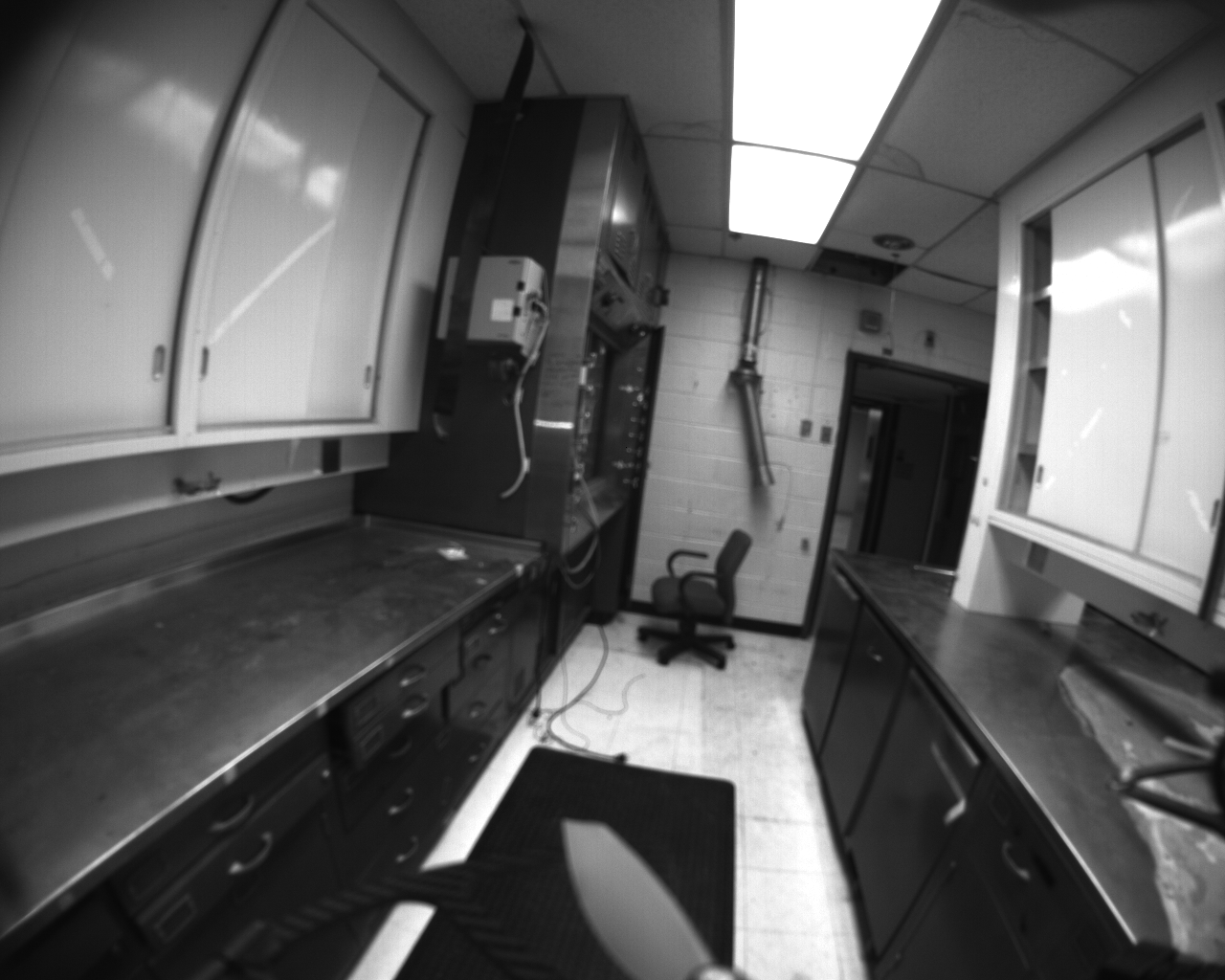}\includegraphics[width=0.25\linewidth]{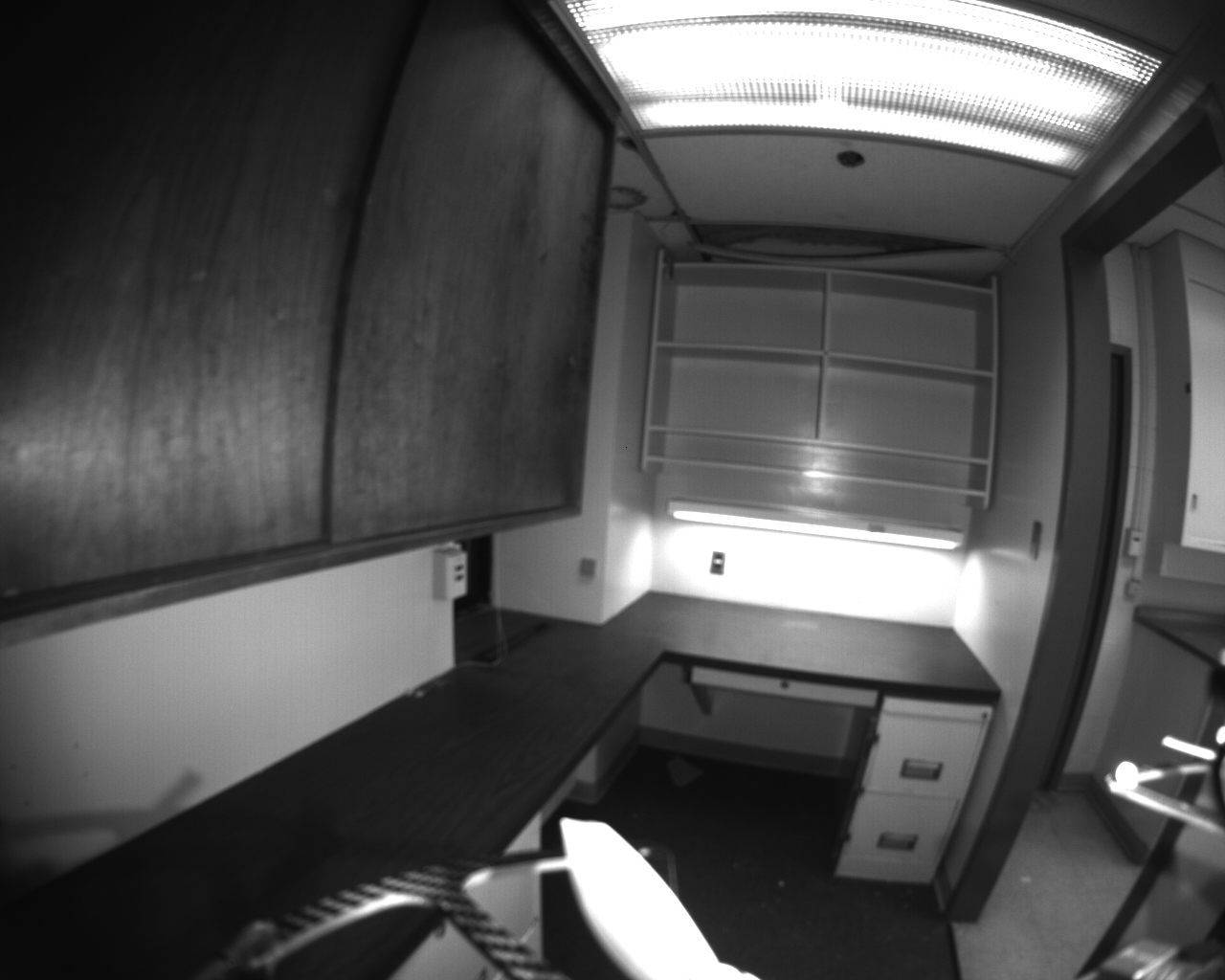}
        \caption{Sample Images}
        \label{fig:online2snapshots}
    \end{subfigure}
    \begin{subfigure}{\textwidth}
        \centering
        \includegraphics[width=0.12\linewidth,height=1.5in,keepaspectratio,clip,trim={4.8cm 0cm 5.7cm 1.5cm}]{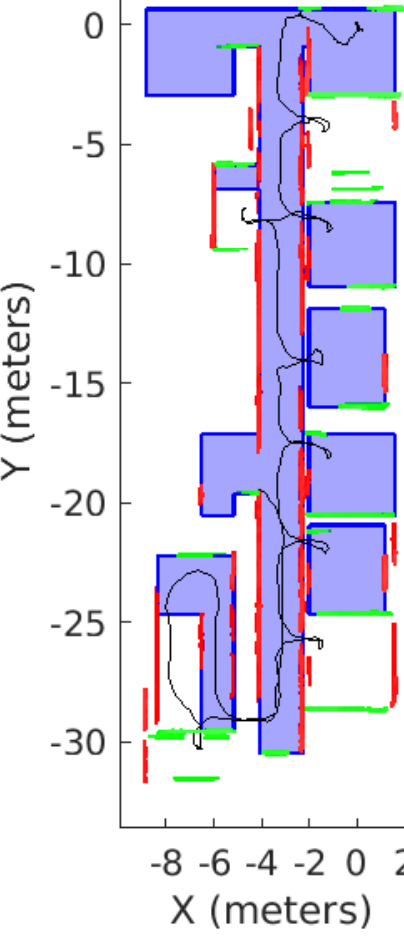}\quad \includegraphics[width=0.15\linewidth,height=1.5in,keepaspectratio,clip,trim={3.8cm 0cm 4.5cm 0.9cm}]{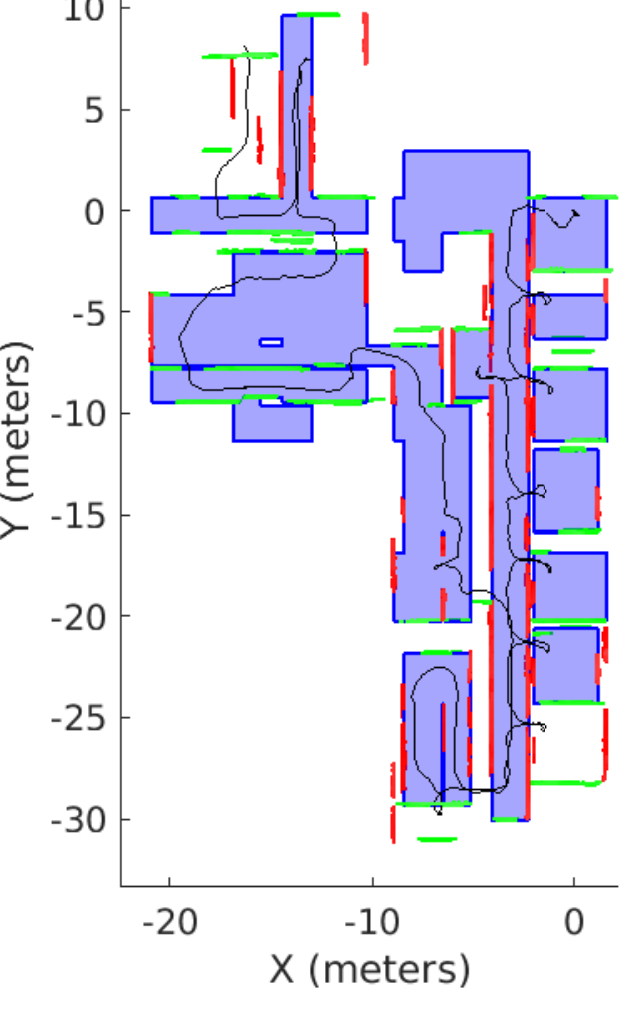}\quad\includegraphics[width=0.35\linewidth,height=2in,keepaspectratio,clip,trim={0.5cm 1.9cm 1.4cm 2.6cm}]{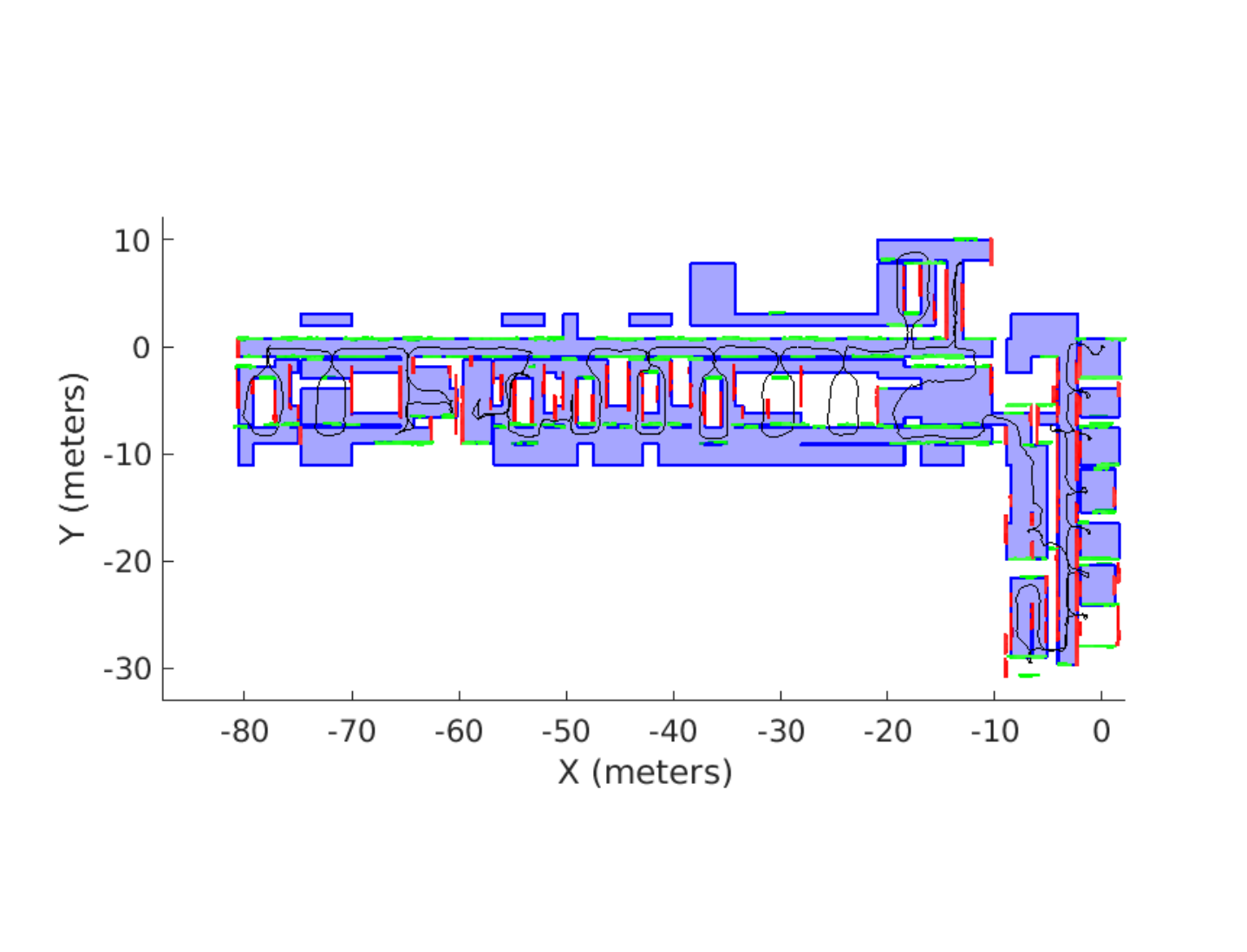}\quad\includegraphics[width=0.35\linewidth,height=2in,keepaspectratio,clip,trim={0.6cm 2.1cm 2.1cm 3cm}]{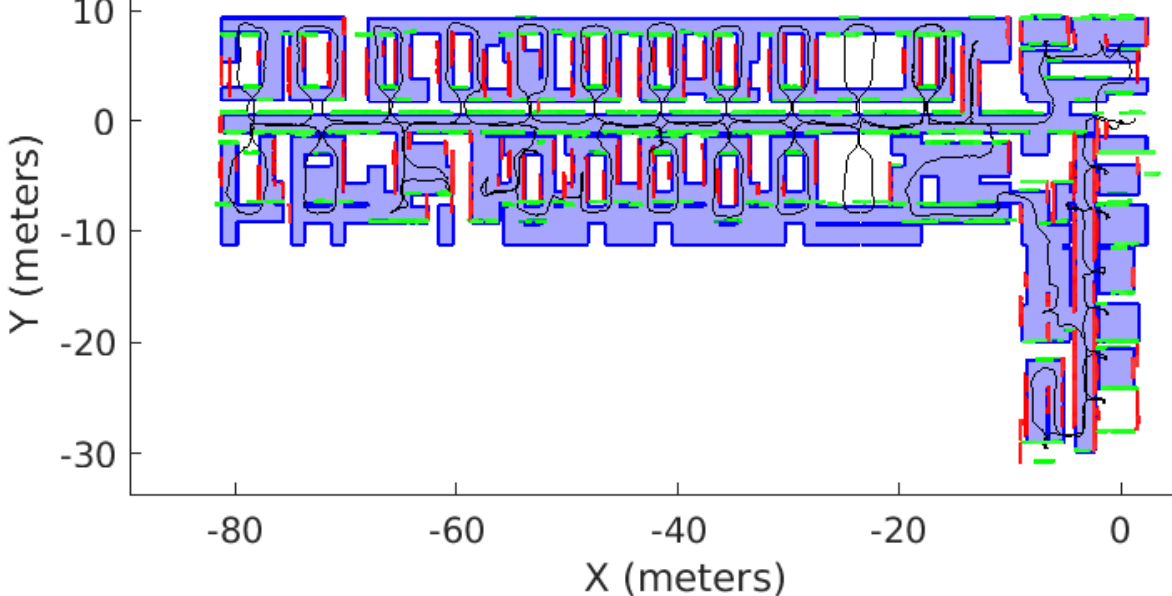}
        \caption{Estimated floor plan at $t=600$, $1000$, $2400$, and $3926$ seconds.}
        \label{fig:online2}
    \end{subfigure}
    \caption{Online estimation results for Building B. Total distance traveled $=767$ meters. }
    \label{fig:my_label2}
\end{figure*}

The procedure was carried out in two extended environments and the results are shown in Figures \ref{fig:online1} and \ref{fig:online2}. Although both spaces are of roughly the same dimension, $80 \times 40$ meters, the length of the exploratory path taken through these spaces as well as their respective topologies are quite different. Sample images taken in both environments can be seen in Figures \ref{fig:online1images} and \ref{fig:online2snapshots}, which provide more context for the types of environments being explored.

The first environment is an academic building featuring vast hallways, large classrooms, and highly visible walls. These qualities naturally lead to simpler space, which leads to a faster convergence and a more accurate model being produced. The second environment is an abandoned industrial laboratory and contains a larger number of densely packed interconnected rooms with more built in furniture which occludes the structural wall surfaces. This more complex structure results in fewer observations of the dominant structure, but a significant overall increase in the total number of planes detected, and as a result leads to a more challenging optimization. The sequence is also significantly longer than first ($767$ meters vs $247$ meters). Nonetheless, despite these challenges the system is still able to extract the major structural features of the space and produce an estimate for the floor plan. 

Again, as the exploration proceeds and the system learns about more structural planes it is able to use these to posit more accurate completions of the space. For example the system is able to apprehend the dimensions of neighboring rooms on a corridor by suggesting that they share some of the same structural walls even when those surfaces are not directly observed in each room since the optimization algorithm favors simple, regular explanations.

\section{Conclusion}


In this paper we have described an algorithm that can be used to automatically extract plausible floor plans of indoor scenes based
on the kinds of incomplete 3D data that an autonomous mobile robot could acquire. The extracted floor plan is designed to be useful for subsequent motion planning procedures since it produces water tight explanations of space that complete partially observed layout surfaces and infer likely regions of free space. This ability to better understand the space from limited data allows the system to construct better motion plans with less data obviating the need to exhaustively explore each corner of the scene.

Our approach exploits an algorithm that provides it with estimates for the salient structural planes in the scene and it constructs volumetric explanations using those infinite planes as boundaries. Importantly, this allows the system to suggest relationships between rooms that are not evident in the acquired data. The algorithm also exploits the connection between the layout estimation task and the set cover problem to propose effective optimization algorithms with provable performance guarantees.

The ability to accurately extract room layout structure is an important first step in scene interpretation. These results provide context which can be used to inform other semantic analysis operations such as detecting and positioning furniture and speculating about the function of different spaces. Ultimately, it helps the autonomous system to apprehend the scene at a higher level of abstraction and communicate more effectively with human interlocutors.


\clearpage

\clearpage

{\small
\bibliographystyle{ieee}
\bibliography{Layout_Estimation,Floor_Plan_Generation,misc}
}

\end{document}